\documentclass[runningheads]{llncs}

\usepackage{eccv}
\usepackage[width=122mm,left=12mm,paperwidth=146mm,height=193mm,top=12mm,paperheight=217mm]{geometry}

\usepackage{eccvabbrv}

\usepackage{graphicx}
\usepackage{booktabs}

\usepackage[accsupp]{axessibility}  

\usepackage{adjustbox}
\usepackage{comment}
\usepackage{physics}
\usepackage{multirow}

\usepackage{todonotes}

\usepackage{hyperref}

\usepackage{orcidlink}

\begin{document}

\title{Camera Height Doesn't Change:\\
Unsupervised Training for Metric Monocular Road-Scene Depth Estimation} 

\titlerunning{Unsupervised Training for Metric Monocular Road-Scene Depth Estimation}

\author{Genki Kinoshita\orcidlink{0009-0005-9076-3152} \and
Ko Nishino\orcidlink{0000-0002-3534-3447}}

\authorrunning{G.Kinoshita et al.}

\institute{Graduate School of Informatics, Kyoto University, Kyoto, Japan
\email{gkinoshita@vision.ist.i.kyoto-u.ac.jp, kon@i.kyoto-u.ac.jp}\\
\url{https://vision.ist.i.kyoto-u.ac.jp/}}

\maketitle

\definecolor{mygray}{gray}{0.7}
\newcommand{\gray}[1]{\textcolor{mygray}{#1}}
\newcommand{\red}[1]{\textcolor{red}{#1}}
\newcommand{\cam}{\mathrm{cam}}
\newcommand{\obj}{\mathrm{obj}}
\newcommand{\aux}{\mathrm{aux}}
\newcommand{\road}{\mathrm{road}}
\newcommand{\depth}{\mathrm{depth}}
\newcommand{\pose}{\mathrm{pose}}
\newcommand{\rec}{\mathrm{rec}}
\newcommand{\photo}{\mathrm{photo}}
\newcommand{\sm}{\mathrm{sm}}
\newcommand{\SSIM}{\mathrm{SSIM}}
\newcommand{\tgt}{\mathrm{tgt}}
\newcommand{\src}{\mathrm{src}}
\newcommand{\loss}{\mathcal{L}}

\begin{abstract}
In this paper, we introduce a novel training method for making any monocular depth network learn absolute scale and estimate metric road-scene depth just from regular training data, \ie, driving videos. We refer to this training framework as FUMET. The key idea is to leverage cars found on the road as sources of scale supervision and to incorporate them in network training robustly. FUMET detects and estimates the sizes of cars in a frame and aggregates scale information extracted from them into an estimate of the camera height whose consistency across the entire video sequence is enforced as scale supervision. This realizes robust unsupervised training of any, otherwise scale-oblivious, monocular depth network so that they become not only scale-aware but also metric-accurate without the need for auxiliary sensors and extra supervision. Extensive experiments on the KITTI and the Cityscapes datasets show the effectiveness of FUMET, which achieves state-of-the-art accuracy. We also show that FUMET enables training on mixed datasets of different camera heights, which leads to larger-scale training and better generalization. Metric depth reconstruction is essential in any road-scene visual modeling, and FUMET democratizes its deployment by establishing the means to convert any model into a metric depth estimator.

\keywords{Unsupervised Monocular Depth Estimation \and Metric Scale \and Camera Height Invariance \and Object Size Prior}
\end{abstract}

\section{Introduction}
\label{sec:intro}

Monocular depth estimation is essential for autonomous driving and advanced driver assistance systems (ADAS). It underlies many of their key perceptual tasks including 3D object detection~\cite{D4LCN, Pseudo-LiDAR, Pseudo-LiDAR++, AreWeMissing}, motion planning~\cite{MotionPlanning, NeuralMotionPlanner, SafeMotionPlanning}, and road segmentation~\cite{RoadSegm}. Although supervised methods have achieved impressive accuracy, they suffer from costly ground-truth data collection. This limits their generalizability~\cite{SlowTV} and also makes continuous learning difficult.

Many recent methods~\cite{FreqAware, Lite-Mono, HR-Depth, Monodepth2, RA-Depth} leverage self-supervision to scale up training. Most of these methods, however, suffer from scale ambiguity which renders them difficult to use off-the-shelf. To resolve the scale ambiguity and recover in metric scale, the actual scene scale needs to be supervised in some form. Past works have leveraged velocity~\cite{PackNet}, GPS~\cite{GPS}, IMU~\cite{IMU, IMUICRA}, or camera height~\cite{ScaleRecovery, VADepth} for this. These methods require auxiliary sensors to measure these extra quantities in addition to the monocular RGB camera, which significantly limits the ability to scale the size of training data. Arbitrary in-the-wild driving videos cannot be used as they lack such measurements. These self-supervised methods often still require weak scale supervision and cannot be made fully self-supervised as the scale changes for each sequence captured with a different camera. 

Road scenes actually have plenty of untapped sources of metric scale. Cars found on the road are one of them as they are rigid objects whose actual sizes do not change and are unique to each make and model. The road is also exactly the region where we need accurate metric depth the most. If we can leverage the sizes of cars to bridge the 3D to 2D projection, we may train a mono-depth network to become metric-accurate. Simply employing a prior on the car size, however, would be too brittle since the metric supervision will be as ambiguous as the accuracy of that prior. The perspective projection and the variations in car makes, years, and colors, lead to a large variance in the size estimates, which will directly affect the mono-depth metric accuracy. In addition, there are no reliable object size priors except for a few introduced as probability densities for other tasks~\cite{Metrology, PuttingObjects, MetricTree} whose parameters are not shared.

How can we train an arbitrary mono-depth estimator to become metric-accurate by robustly leveraging prior knowledge about the 3D sizes of cars found in driving videos? We derive \textbf{\underline{f}}ully  \textbf{\underline{u}}nsupervised  \textbf{\underline{me}}tric depth \textbf{\underline{t}}rainer (\textbf{FUMET}) to answer this question. Our key idea is to aggregate car size cues in the frames and the sequence into a single physical measure, namely the camera height. Regardless of the scene, the camera capturing it is the same for the same sequence. This means that, by transforming the car size cues found in different frames into a camera height estimate, the very fact that it should not change can be used as a supervision of invariance to achieve metric depth reconstruction. We can formulate this as an optimization of the camera height across frames and training epochs. This leads to gradual metric-aware learning that is stable and accurate, which also makes the depth estimates consistent across the sequence.

We introduce a \textbf{\underline{l}}earned \textbf{\underline{s}}ize \textbf{\underline{p}}rior (\textbf{LSP}) to estimate the size of each car found in a frame. This prior gives us the vehicle dimensions from its appearance. It is trained with various augmentations on a large-scale dataset without the need for any manual annotation. By comparing these dimension estimates with those computed from the depth estimates, we obtain a per-frame scale factor. The multiplication of this scale estimate with the camera height estimate derived from the estimated depth becomes the camera height estimate of the frame. These estimated camera heights across all frames in a sequence are then consolidated by using their median value which is then used as scale supervision with its weighted moving average at the end of each training epoch.

We evaluate the effectiveness of FUMET with extensive experiments using the KITTI and the Cityscapes datasets and compare its accuracy with a number of weakly-supervised methods. The results clearly show that FUMET succeeds in making monocular depth estimation methods learn metric scales and achieves the state-of-the-art accuracy without any direct depth and scale supervision. 
We also demonstrate its capability to train a model on mixed datasets collected with different cameras, \ie, camera heights. This enables larger-scale training and leads to higher generalization. We believe FUMET can serve as a fundamental building block for monocular depth estimation in the wild.


\section{Related work}
\label{sec:RelatedWork}
\subsection{Self-Supervision}
Although, as in other tasks, supervised methods achieve the highest accuracy in monocular depth estimation (MDE)~\cite{AdaBins, TrapAttention, GLPDepth, NeWCRF, iDisc}, they obviously require ground-truth data usually obtained with a LiDAR sensor. 
Self-supervised MDE methods remove the need for such hardware costs for training. Garg \etal~\cite{FirstUnsupervised} proposed the first approach to self-supervised MDE by predicting per-pixel disparities between a stereo pair. Godard \etal~\cite{Monodepth} extend this idea by introducing left-right consistency and greatly improve the accuracy. These methods require calibrated stereo data for training which limits their applicability.
Zhou \etal~\cite{SfMLearner} introduced a pioneering method for training solely from monocular videos by incorporating a network that estimates the relative camera poses between successive frames. Based on this idea, many MDE methods that can be trained solely with monocular videos have been proposed~\cite{Monodepth2, HR-Depth, DIFFNet, Lite-Mono, FreqAware, RA-Depth}.

Despite the popularity and improvements of self-supervised MDE, directly deploying those models in autonomous driving or ADAS systems is unrealistic as they only produce relative depth up to an unknown scale. 
Our FUMET overcomes this issue without the need for any external supervision. 

\subsection{Weak Supervision}
Several works have tackled scale-aware MDE by adopting weak scale supervision through velocity~\cite{PackNet}, GPS~\cite{GPS}, IMU~\cite{IMUICRA}, or camera height~\cite{ScaleRecovery, VADepth} measurements. 
Zhang \etal~\cite{IMU} use the combination of IMU, velocity, and gravity to achieve the highest accuracy among these methods. 
These methods fundamentally rely on such auxiliary measurements in addition to regular RGB images. This makes it impossible to take advantage of the many road-scene videos on the Internet (\eg, YouTube) or casually installed dashcam videos for training. 
In contrast, FUMET relies on a universal car size prior which is independent of the data collection environments and does not require any such weak supervision; it can be trained solely from monocular video sequences.

\subsection{Object Size Priors}
Several works for other tasks leverage priors on object sizes including single view metrology~\cite{Metrology}, object insertion~\cite{PuttingObjects, MetricTree}, and visual odometry~\cite{SLAM, BayesianSLAM}. These methods model the priors as normal distributions or Gaussian mixture models, and the parameters of these distributions are predetermined. When there is a gap between those values and the true ones, which is usually the case in the real world, the difference directly surfaces in erroneous scale estimates. 

Instead of these fixed priors, we introduce a learning-based size prior which we refer to as LSP. This estimates vehicle dimensions based on its appearance, thus can avoid the aforementioned problem. Furthermore, it is trained on a large-scale dataset with various augmentations, so that it can robustly estimate vehicle dimensions from any image.


\section{Preliminaries}
Our goal is to train a metric-accurate MDE model $\theta_\depth$ with only self-super-vision.
This self-supervision is comprised of two parts: one designed for learning metric scale (\cref{sec:FUMET}) and the other dedicated to learning depth geometry. For the latter, we follow Zhou \etal~\cite{SfMLearner} and minimize the reconstruction loss between the current frame $I_t$ and the ones synthesized from temporally adjacent frames $I_s,(s=t\pm1)$.
We employ a pose network $\theta_\pose$ in addition to $\theta_\depth$ to estimate the relative camera pose $\vb{T}_{t\rightarrow s} \in \mathrm{SE}(3)$ and the current dense depth map $D_t$, respectively, 
\begin{equation}
    D_t = \theta_\depth(I_t),\,\,\vb{T}_{t\rightarrow s} = \theta_\pose(I_t, I_s)\,.
\end{equation}
We then synthesize the image from the current viewpoint 
\begin{align}
    I_{s\rightarrow t} = I_s \ev{proj(D_t, \vb{T}_{t\rightarrow s}, \vb{K})}\,,
\end{align}
where $\ev{\cdot}$ is a bilinear sampling operator, $proj(\cdot)$ outputs the 2D coordinates of $D_t$ when projected into the viewpoint of $I_s$, and $\vb{K}$ is the camera intrinsics. 
Following Godard \etal~\cite{Monodepth2}, we use an affine combination of $\SSIM$~\cite{SSIM} and L1 loss as the photometric error $pe$ and define the reconstruction loss $\loss_\rec$ as the photometric error between $I_t$ and $I_{s\rightarrow t}$ for the pixels with minimum error across source frames
\begin{align}
   \label{eq:RecLoss}
   &\loss_\rec = \min_s pe(I_t, I_{s\rightarrow t}) \,,  \\
   &pe(I_t, I_{s\rightarrow t}) = \frac{\lambda}{2}(1 - \SSIM(I_t, I_{s\rightarrow t})) + (1 - \lambda) |I_t - I_{s\rightarrow t}|\,,
\end{align}
where $\lambda=0.85$. We also employ an edge-aware smoothness loss~\cite{Monodepth}
\begin{align}
    \loss_\sm = | \partial_x d^*_t | e^{-|\partial_x I_t|} + | \partial_y d^*_t | e^{-|\partial_y I_t|}\,,
\end{align}
where $d^*_t = d_t \big/\bar{d_t}$ is the mean-normalized inverse depth~\cite{DDVO} to prevent the estimated depth from shrinking.
Furthermore, when computing $\loss_\rec$ we use auto-masking~\cite{Monodepth2} to mask out stationary pixels which generate incorrect supervisory signals.


\begin{figure*}
  \centering
  \includegraphics[width=\linewidth]{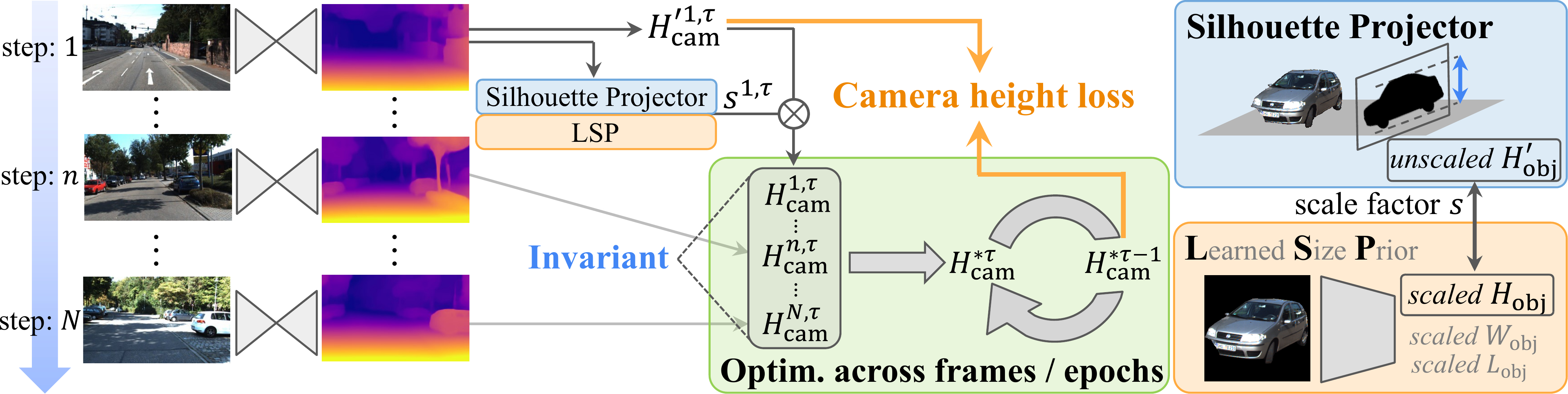}
  \caption{Overview of FUMET. At each training step $n$, an \emph{unscaled} camera height $H'^{n,\tau}_{\cam}$ is computed differentiablly from the estimated depth. The previous epoch $\tau -1$ provides supervision with a scaled camera height $H^{*\tau-1}_{\cam}$. To obtain this scaled camera height supervision, Silhouette Projector first computes the object silhouette heights $H'_\obj$ from the depth map. 
  By comparing $H'_\obj$ and the estimated one $H_\obj$ with LSP, per-frame scale factor $s$ is determined and we obtain the scaled camera height $H^{n,\tau}_\cam = s\cdot H'^{n,\tau}_\cam$.
  At the end of each epoch, $H^*_\cam$ is optimized across a series of consecutive frames and updated with the weighted moving average.}
  \label{fig:Overview}
\end{figure*}

\section{FUMET}
\label{sec:FUMET}
\cref{fig:Overview} depicts the overall architecture of FUMET. Please see the caption for an overview of the framework. Let ``$\,'\,$'' denote ``not metrically scaled''.

\subsection{Scale-Aware Self-Supervised Learning}

In FUMET, we derive a camera height pseudo-supervision from depth estimates. To obtain the per-frame camera height, we first compute the normal vector for each pixel $p_i$.
We consider the 8-neighborhood of $p_i$ and form 8 pairs $\mathcal{N}(p_i) = \{ (p^i_{j_0}, p^i_{j_1}) \}^8_{j=1}$. Each pair $(p^i_{j_0}, p^i_{j_1})$ consists of two pixels whose difference vectors w.r.t. $p_i$ are perpendicular to each other and are sorted counterclockwise. We reproject the paired pixels onto 3D space and compute the cross-product of the two difference vectors w.r.t. the reprojected 3D point of $p_i$. By summing and normalizing all the cross-product vectors of each pair, we obtain the normal vector $\hat{\vb{n}}(p_i)$
\begin{equation}
    \label{eq:normal}
    \vb{n}(p_i) = \!\!\sum_{\mathcal{N}(p_i)} (\vb*{\phi}(p^i_{j_0})\! - \vb*{\phi}(p_i)) \times (\vb*{\phi}(p^i_{j_1})\! - \vb*{\phi}(p_i))\,,\,\,\, \hat{\vb{n}}(p_i) = \frac{\vb{n}(p_i)}{\|\vb{n}(p_i)\|}\,,
\end{equation}
where $\vb*{\phi}(\cdot)$ denotes perspective reprojection. 

For pixels in the road region, the negated inner product between the normal vector and the reprojected 3D point corresponds to the camera height from the ground plane
\begin{equation}
    H'_\cam(p_i) = -\,\vb*{\phi}(p_i) \cdot \hat{\vb{n}}(p_i)\,.
\end{equation}
We represent the median of $H'_\cam(p_i)$ as the camera height of the frame 
\begin{equation}
    \tilde{H}'_\cam = \mathrm{med}(\{H'_\cam(p_i) \mid p_i\in M_\mathrm{r}\})\,,
\end{equation}
where $p_i$ is in the road region mask $M_\mathrm{r}$ obtained with an off-the-shelf semantic segmentation model~\cite{OneFormer} before training. This camera height $\tilde{H}'_\cam$ estimate is \emph{unscaled} and needs to be upgraded to a real-scaled value to impose the invariance loss across frames. 
By using the Silhouette Projector, we aggregate the scale information from the object height prior into a per-frame scale factor $s$ (see \cref{subsec:SilhouetteProjector}) and obtain the \emph{scaled} camera height as $H_\cam=s \cdot \tilde{H}'_\cam$. 

We leverage the fact that the camera height does not change in a sequence and formulate the invariance into an optimization across frames and training epochs. Thanks to this optimization, we acquire an accurate and stable supervision, which is also consistent across the scene. 
At the end of the $\tau$-th epoch, we use the median of $\{H^{n,\tau}_\cam\}^N_{n=1}$ for all frames $N$ as the representative camera height $H^\tau_\cam$ at the $\tau$-th epoch. Then, we update the \emph{scaled} camera height pseudo-invariance $H^*_\cam$ with its weighted moving average
\begin{align}
    \label{eq:UpdateCamHeight}
    H^{*\tau}_\cam \leftarrow \bigg\{ \frac{\tau(\tau-1)}{2} \cdot H^{*\tau-1}_\cam + \tau \cdot H^\tau_\cam \bigg\} \bigg/ \frac{\tau(\tau+1)}{2}\,.
\end{align}
When training on a dataset containing images captured at multiple camera heights, we create pseudo-labels corresponding to each camera height and optimize them individually.

To make the network learn the absolute scale and directly produce the \emph{scaled} depth map, we enforce a loss on the camera height $H'_\cam$ using the pseudo-invariance  $H^{*\tau-1}_\cam$ computed in the previous epoch 
\begin{equation}
    \loss_\cam = \frac{1}{| M_\mathrm{r} |} \sum_i M_\mathrm{r}(p_i) \cdot | H'_\cam(p_i) - H^{*\tau-1}_\cam|\,.
\end{equation}

It is worth noting that at inference, the network does not assume camera height invariance in the input sequence; it is only assumed during training. The computational cost at inference is exactly the same as the original MDE model trained without FUMET.

\subsection{Silhouette Projector}
\label{subsec:SilhouetteProjector}

In the process of creating a \emph{scaled} camera height supervision, we need to know the per-frame scale factor. For this, we can consider leveraging an object of known size. This, however, would necessitate extraction of the exact 3D object region which is non-trivial as objects are often occluded or cropped and oriented in various poses. 

We introduce Silhouette Projector to robustly estimate this per-frame scale factor. Silhouette Projector leverages two simple facts of road scenes. First, the height of the object silhouette projected on the plane perpendicular to the ground plane does not change regardless of the object pose. Second, we can compute the silhouette height by measuring the distance from the top of the silhouette to the ground plane even if they are partly occluded or truncated, as long as the top is visible and the object is on the ground plane.

We first reconstruct an object point cloud with the estimated depth map through perspective reprojection. The object region $M_\obj$ in the image can be obtained with an off-the-shelf instance segmentation model~\cite{OneFormer} before training.
We then orthographically project these 3D points onto a plane $P^{\perp}$ perpendicular to the ground plane, \ie, an arbitrary plane parallel to the road normal vector $\vb{\tilde{n}}$. We compute $\vb{\tilde{n}}$ as the median of the normal vectors $\hat{\vb{n}}(p_i)$ (see \cref{eq:normal}) in the road region $M_\mathrm{r}$. We define the farthest distance from the 2D point on $P^{\perp}$ to the ground plane as the silhouette height. This value is \emph{unscaled} because we derived this from the estimated \emph{unscaled} depth map. 
The ratio between a height prior and its silhouette height corresponds to the scale factor of interest.
When there are multiple objects in an image, we use the median of the scale factors each corresponding to each object as the scale factor of the frame for computational simplicity and robustness.

\subsubsection{Outlier Filtering.}
Some objects cannot be handled well with Silhouette Projector for various reasons including occluded upper parts, not on the ground plane, inaccurate segmented region, and misclassified object category. We introduce an algorithm to detect these as outliers based on geometric plausibility. 
We first calculate the horizon $\vb{l}$ with the road normal vector $\vb{\tilde{n}}$ and the camera intrinsics $\vb{K}$~\cite{Hartley}
\begin{equation}
    \vb{l} = \vb{K}^{-\mathsf{T}}\vb{\tilde{n}}\,.
\end{equation}
Based on the approximations in~\cite{PuttingObjects}, we derive the approximate metric-scale object height $\hat{H}_\obj \approx \frac{h_\obj}{h_\cam} H^{*\tau-1}_\cam$, 
where $h_\cam$ is the farthest distance from the pixel in the object instance mask to $\vb{l}$, $h_\obj$ is the farthest distance between two points on the object mask on which tangent lines parallel to $\vb{l}$ lie, and $H^{*\tau-1}_\cam$ is the pseudo camera height at the $\tau$-th epoch derived in \cref{eq:UpdateCamHeight}. By comparing the relative gap between $\hat{H}_\obj$ and the object height prior $H_\obj$, we determine the outliers as
\begin{equation}
    \bigg\{ k \,\bigg| \frac{| H_{\obj}^k - \hat{H}_{\obj}^k |}{H_{\obj}^k}  > \mathcal{T} \bigg\}\,.
\end{equation}
In all the experiments, we set the threshold $\mathcal{T}$ to 0.2.

\subsection{Training Losses}
\subsubsection{Auxiliary Rough Geometric Loss.} 
We experimentally observe that estimated object depths, especially those distant from the camera, often erroneously lean towards the road surface because of the inaccurate supervisory signal from the reconstruction loss in low-texture regions.
This tendency makes silhouette heights inaccurate. To mitigate this issue, we introduce an auxiliary rough geometric loss $\loss_\aux$, which enforces object point clouds to be within a close region. 
For each inlier object $k\,(k=1 \dots K)$, we define $\loss_\aux$ as the gap between the estimated depth map $D$ in the object region $M_\obj$ and the approximate object depth $D_\mathrm{aprx} = \frac{H_\obj}{h_\obj}f_y$ where $f_y$ is the focal length
\begin{equation}
    \loss_\aux \!=\! \frac{1}{K} \sum_{k} \frac{1}{|M^k_\obj|} \!\sum_i M^k_\obj(p_i) \cdot \bigg| D(p_i) - D^k_{\mathrm{aprx}}  \bigg|\,.
\end{equation}
As a by-product, this loss accelerates learning the metric scale.

The $\loss_\aux$ is potentially inaccurate since $D_\mathrm{aprx}$ assumes that each object can be represented by a plane parallel to the image plane, and we observe that it degrades the depth accuracy in the late stage of training. On the other hand, in the early stage, estimated depth is unreliable, so heavily relying on the camera height loss $\loss_\cam$ makes the training unstable. For this, we gradually decrease the weight of $\loss_\aux$ and increase the one for $\loss_\cam$ as the training proceeds in a logarithmic way. To ensure training stability, we stop changing these weights after the $\tau_\mathrm{mid}$-th epoch 
\begin{equation}
    \lambda_\aux(\tau) = 
        \begin{cases}
            \frac{-\log(\tau + 1)}{\log(\tau_\mathrm{mid} + 1)} & \!\!\!(\tau \leq \tau_\mathrm{mid}) \\
            \epsilon & \!\!\!(\tau > \tau_\mathrm{mid})
        \end{cases},\,\,
    \lambda_\cam(\tau) = 
        \begin{cases}
            \frac{\log(\tau + 1)}{\log(\tau_\mathrm{mid} + 1)} & \!\!\!(\tau \leq \tau_\mathrm{mid}) \\
            1.0 & \!\!\!(\tau > \tau_\mathrm{mid})
        \end{cases},
    \label{eq:BalanceWeight}
\end{equation}
where $\epsilon=0.005$.

The total loss is $\loss = \alpha\lambda_\cam\loss_\cam + \beta\lambda_\aux\loss_\aux + \loss_\sm + \loss_\rec$ where $\alpha=0.01$ and $\beta=1.0$.


\section{LSP}
\label{sec:LearnedSizePrior}

Our FUMET extracts metric scale from a prior on object heights through the camera height loss $\loss_\cam$ and the auxiliary rough geometric loss $\loss_\aux$.
To ensure that FUMET learns the scale regardless of the particular dataset used for training, a universal prior of object heights, independent of the environment, becomes essential. 
Cars are objects whose dimensions are explicitly and accurately known. Better yet, they are abundantly located on the road, the key semantic region whose metric depth is essential for many downstream tasks.  
Some methods for other tasks have modeled the size prior as a probability distribution~\cite{PuttingObjects, Metrology}. The gap between its mean and the true expectation for each environment, however, is directly reflected in the uncertainty of the scale in the supervision. Wang \etal~\cite{TrainInGermany} also report that there is a gap in car dimensions among datasets. These facts suggest that modeling the object size prior as a single probability distribution is problematic, especially for man-made objects like cars. 

We humans not only possess rough prior knowledge about the vehicle size but can also estimate it more accurately by extracting instance-specific information such as car models from its appearance. Motivated by this intuition, we introduce a learned universal vehicle size prior termed LSP. Given a masked vehicle image, LSP estimates its dimensions.

For training, we use a dataset consisting of a large number of vehicle images and their information such as 3D dimensions. This dataset can be easily collected with web scraping and does not require manual annotations.
To enhance the robustness to various input images, in addition to common color jittering and blur augmentation, we perform augmentations to simulate occlusion and truncation.
For occlusion augmentation, we mask out random regions with the same shape as a vehicle or with a beam of random width. For truncation augmentation, we translate the vehicle mask in a random direction. We visualize these in the supplementary material.
The abundance of training images and diverse augmentations enable LSP to estimate vehicle dimensions robustly across various datasets.

Although FUMET requires only a car height prior, we construct LSP to estimate the width and length as well as height. This enhances it to grasp an overall car size and improves the prediction accuracy of car heights.

\section{Experiments}

\subsection{FUMET}
\label{sec:FUMETResults}

We evaluate the accuracy of depth estimators trained on our framework and demonstrate the effectiveness of FUMET as a novel unsupervised metric-scale training method through depth prediction accuracy, its applicability to any network architecture, and its ability to leverage any dataset for training. 

\subsubsection{Implementation Details.}
\label{sec:ImplementationDetails}

We implement FUMET with PyTorch~\cite{PyTorch}. We employ Monodepth2~\cite{Monodepth2} with ResNet50 encoder pre-trained on ImageNet~\cite{ImageNet} as the default depth estimator which is one of the most basic architectures in self-supervised MDE. For all experiments unless mentioned otherwise, we train the model for 50 epochs with a batch size of 8 and set $\tau_\mathrm{mid}$ to 20. We use Adam optimizer~\cite{Adam} with an initial learning rate of $5.0 \times 10^{-5}$ decayed by half every 15 epochs.
During training, we perform the following augmentations with 50\% chance in random order: horizontal flips, brightness adjustment ($\pm$0.2), saturation adjustment ($\pm$0.2), contrast adjustment ($\pm$0.2), and hue jitter ($\pm$0.1). 
For quantitative evaluation, we follow~\cite{StandardMetrics} and compute the seven standard metrics: AbsRel, SqRel, RMSE, $\mathrm{RMSE_{log}}$, $\delta<1.25$, $\delta<1.25^2$, and $\delta<1.25^3$.

\begin{table*}[t!]
    \caption{Quantitative results on the KITTI Eigen test split~\cite{Eigen} (640$\times$192 resolution). In the Supervision column, V, G, and CamH represent velocity, gravity, and camera height, respectively. The Scaling column represents whether to perform the median scaling~\cite{SfMLearner}.
    We denote the best results in \textbf{bold} and the second best ones with an \underline{underline} for each block. Note that we train VADepth~\cite{VADepth} from scratch using ground-truth camera height (1.65m) provided by KITTI, unlike its original paper. FUMET outperforms weakly-supervised methods regardless of the network architectures and without the need for ground truth supervision. In the rescaled results, every model trained with FUMET achieves better results than its original one.}
    \label{tab:KITTIQuantity}
    \centering
    {
    \scriptsize
    \setlength{\tabcolsep}{0.3em}
    \begin{tabular}{@{}c|c|c|cccc|ccc@{}}
    \toprule

                              &              &                       &  \multicolumn{4}{c|}{Error ($\downarrow$) } & \multicolumn{3}{c}{Accuracy ($\uparrow$)} \\
    Method                    & \!Supervision\!  & \!Scaling\!          & \!AbsRel\!\!            & \!SqRel\!\!             & \!\!RMSE\!\!              & \!RMSE$_{\mathrm{log}}$\! & \!$\delta\!\!<\!\!1.25$\!\!     & \!$\delta\!\!<\!\!1.25^2$\!\!   & \!$\delta\!\!<\!\!1.25^3$\! \\
    \midrule
    
     G2S~\cite{GPS}                            & GPS & - & 0.109 & 0.860 & 4.855 & 0.198 & 0.865 & 0.954 & 0.980 \\
     PackNet-SfM~\cite{PackNet}                  & V & - & 0.111 & 0.829 & 4.788 & 0.199 & 0.864 & 0.954 & 0.980 \\
     Wagstaff \etal~\cite{ScaleRecovery}      & CamH & - & 0.123 & 0.996 & 5.253 & 0.213 & 0.840 & 0.947 & 0.978 \\
     VADepth~\cite{VADepth}                   & CamH & - & 0.120 & 0.975 & 4.971 & 0.203 & 0.867 & 0.956 & 0.979 \\
     DynaDepth~\cite{IMU}              & IMU$+$V$+$G & - & \underline{0.109} & \underline{0.787} & 4.705 & 0.195 & 0.869 & 0.958 & \underline{0.981} \\
     \cmidrule{1-10}
     \textbf{Ours}                                        & - & - & \textbf{0.108} & \textbf{0.785} & 4.736 & 0.195 & \underline{0.871} & 0.958 & \underline{0.981} \\
     Lite-Mono~\cite{Lite-Mono} \!\!$+$\!\! \textbf{Ours} & - & - & 0.112 & 0.798 & \underline{4.692} & 0.194 & \underline{0.871} & \underline{0.959} & \underline{0.981} \\
     HR-Depth~\cite{HR-Depth} \!\!$+$\!\! \textbf{Ours}   & - & - & \underline{0.109} & 0.794 & 4.734 & \underline{0.193} & 0.869 & \underline{0.959} & \textbf{0.982} \\
     VADepth~\cite{VADepth} \!\!$+$\!\! \textbf{Ours}     & - & - & \textbf{0.108} & 0.809 & \textbf{4.572} & \textbf{0.185} & \textbf{0.883} & \textbf{0.963} & \textbf{0.982} \\

    \toprule
    \toprule

    G2S~\cite{GPS}                      & GPS     & \checkmark & 0.112 & 0.894 & 4.852 & 0.192 & 0.877 & 0.958 & 0.981             \\
    PackNet-SfM~\cite{PackNet}          & V       & \checkmark & 0.111 & 0.785 & \textbf{4.601} & 0.189 & 0.878 & 0.960 & \underline{0.982} \\
    Wagstaff \etal~\cite{ScaleRecovery} & CamH    & \checkmark & 0.117 & 0.952 & 4.989 & 0.197 & 0.867 & 0.957 & 0.981  \\
    VADepth~\cite{VADepth}    & CamH    & \checkmark               & 0.110 & 0.977 & 4.872 & \underline{0.187} & \textbf{0.889} & \underline{0.962} & 0.981 \\
    DynaDepth~\cite{IMU}                & IMU$+$V$+$G & \checkmark & \underline{0.108} & \underline{0.761} & 4.608 & \underline{0.187} & 0.883 & \underline{0.962} & \underline{0.982} \\
    \textbf{Ours} & -  & \checkmark                                & \textbf{0.106} & \textbf{0.759} & \underline{4.602} & \textbf{0.184} & \underline{0.887} & \textbf{0.963} & \textbf{0.983}                \\
     
    \toprule
    \toprule
    
    Monodepth2~\cite{Monodepth2} & - & \checkmark & 0.115 & 0.903 & 4.863 & 0.193 & 0.877 & 0.959 & 0.981 \\
    $+$\! \textbf{Ours} & - & \checkmark          & \textbf{0.106} & \textbf{0.759} & \textbf{4.602} & \textbf{0.184} & \textbf{0.887} & \textbf{0.963} & \textbf{0.983} \\
    
    \cmidrule{1-10}
    
    HR-Depth~\cite{HR-Depth} & - & \checkmark   & 0.109 & 0.792 & 4.632 & 0.185 & 0.884 & 0.962 & \textbf{0.983} \\
    $+$\! \textbf{Ours} & - & \checkmark        & \textbf{0.108} & \textbf{0.759} & \textbf{4.584} & \textbf{0.184} & \textbf{0.885} & \textbf{0.963} & \textbf{0.983} \\

    \cmidrule{1-10}
    
    Lite-Mono~\cite{Lite-Mono}   & - & \checkmark & 0.110 & 0.802 & 4.671 & 0.186 & 0.879 & 0.961 & 0.982 \\
    $+$\! \textbf{Ours}   & - & \checkmark        & \textbf{0.107} & \textbf{0.775} & \textbf{4.564} & \textbf{0.183} & \textbf{0.888} & \textbf{0.963} & \textbf{0.983} \\

    \cmidrule{1-10}
    
    VADepth~\cite{VADepth}     & - & \checkmark & 0.104 & 0.774 & 4.552 & 0.181 & 0.892 & 0.965 & 0.982 \\
    $+$\! \textbf{Ours}     & - & \checkmark    & \textbf{0.102} & \textbf{0.760} & \textbf{4.473} & \textbf{0.178} & \textbf{0.897} & \textbf{0.966} & \textbf{0.983} \\
    \bottomrule
    \end{tabular}
    }
\end{table*}

\begin{figure*}[t]
    \centering
    \def\Pa#1{\includegraphics[width=0.235\linewidth]{#1}}
    \begin{tikzpicture}[x=0.99mm,y=1mm]
        \newcount\vX
        \newcount\vY
        \newcount\vdX
        \newcount\vdY
        \newcount\vXi
        \newcount\vdXDyna
        \newcount\vdXGT
        \newcount\vdYGT
        \newcount\vdXErr
        \newcount\vdYErr
        \newcount\gap
        \vdX = 29
        \vdY = -11
        \vXi = -16
        \vdXDyna = 1
        \vdXGT = 12
        \vdYGT = 3
        \vdXErr = 8
        \vdYErr = 3
        \gap = 1

        \vX = \vXi
        \node[inner sep=0pt, font=\tiny] at (\vX,\vY) {\rotatebox[origin=c]{90}{\parbox[t]{20mm}{\centering Input}}};
        \advance\vY by \vdY
        \node[inner sep=0pt, font=\tiny] at (\vX,\vY) {\rotatebox[origin=c]{90}{\parbox[t]{20mm}{\centering \textbf{FUMET}}}};
        \advance\vY by \vdY

        \advance \vX by -\vdXDyna
        \node[inner sep=0pt, font=\tiny] at (\vX,\vY) {\rotatebox[origin=c]{90}{\parbox[t]{20mm}{\centering Dyna\\Depth\!\!~\cite{IMU}}}};
        \advance \vX by \vdXDyna

        \advance\vY by \vdY
        \node[inner sep=0pt, font=\tiny] at (\vX,\vY) {\rotatebox[origin=c]{90}{\parbox[t]{20mm}{\centering G2S\!\!~\cite{GPS}}}};
        \advance\vY by \vdY
        \node[inner sep=0pt, font=\tiny] at (\vX,\vY) {\rotatebox[origin=c]{90}{\parbox[t]{20mm}{\centering VADepth\!\!~\cite{VADepth}}}};
        \advance\vY by \vdY

        \vX = 0
        \vY = 0
        \node at (\vX,\vY) {\Pa{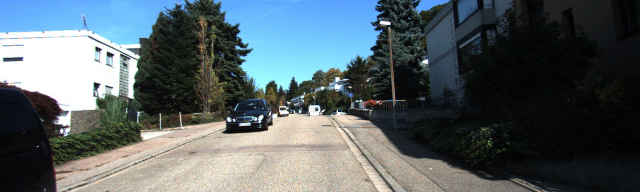}};
        \advance\vY by \vdY
        \node at (\vX,\vY) {\Pa{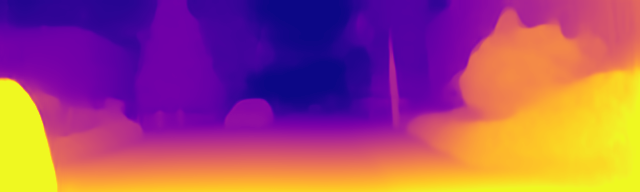}};
        \advance\vY by \vdY
        \node at (\vX,\vY) {\Pa{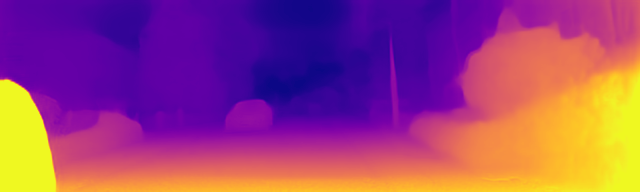}};
        \advance\vY by \vdY
        \node at (\vX,\vY) {\Pa{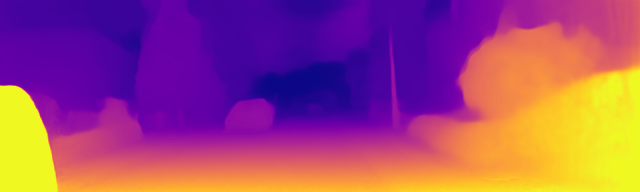}};
        \advance\vY by \vdY
        \node at (\vX,\vY) {\Pa{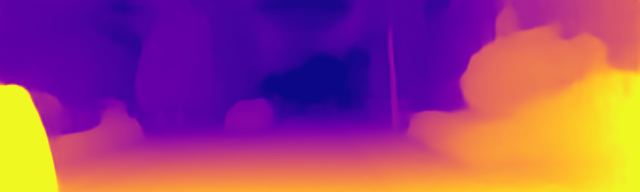}};

        \advance\vX by \vdX
        \vY = 0
        \node at (\vX,\vY) {\Pa{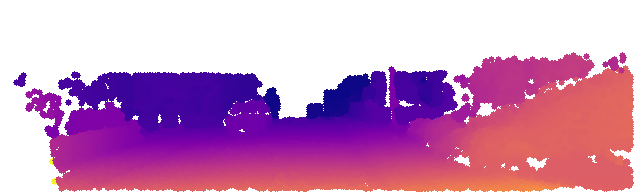}};

        \advance\vX by \vdXGT
        \advance\vY by \vdYGT
        \node[inner sep=0pt, font=\tiny] at (\vX,\vY) {GT};
        \advance\vX by -\vdXGT
        \advance\vY by -\vdYGT

        \advance\vY by \vdY
        \node at (\vX,\vY) {\Pa{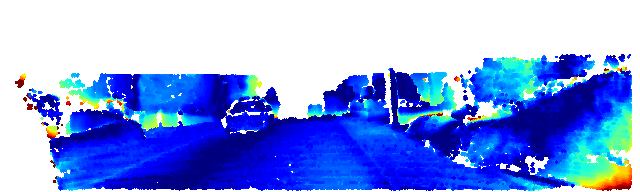}};

        \advance\vX by \vdXErr
        \advance\vY by \vdYErr
        \node[inner sep=0pt, font=\tiny] at (\vX,\vY) {Error map};
        \advance\vX by -\vdXErr
        \advance\vY by -\vdYErr

        \advance\vY by \vdY
        \node at (\vX,\vY) {\Pa{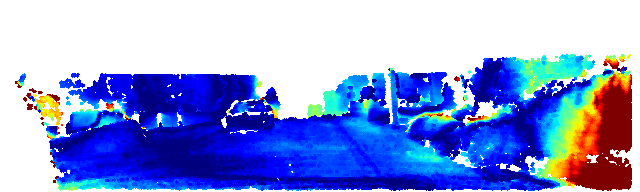}};
        \advance\vY by \vdY
        \node at (\vX,\vY) {\Pa{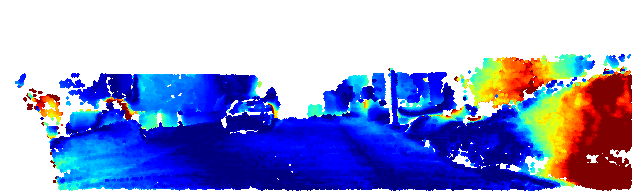}};
        \advance\vY by \vdY
        \node at (\vX,\vY) {\Pa{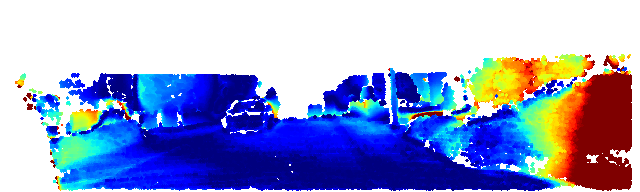}};

        \advance\vX by \vdX
        \advance\vX by \gap
        \vY = 0
        \node at (\vX,\vY) {\Pa{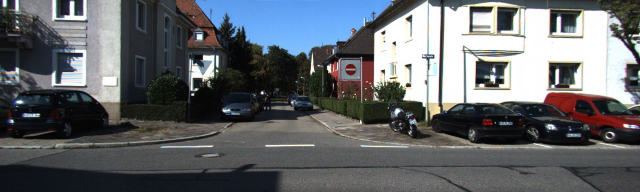}};
        \advance\vY by \vdY
        \node at (\vX,\vY) {\Pa{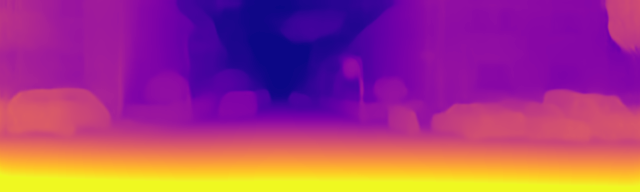}};
        \advance\vY by \vdY
        \node at (\vX,\vY) {\Pa{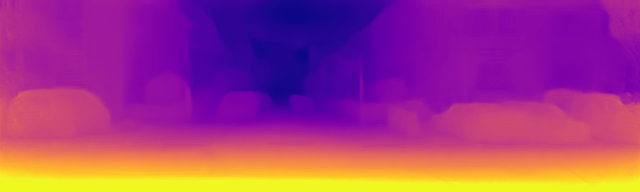}};
        \advance\vY by \vdY
        \node at (\vX,\vY) {\Pa{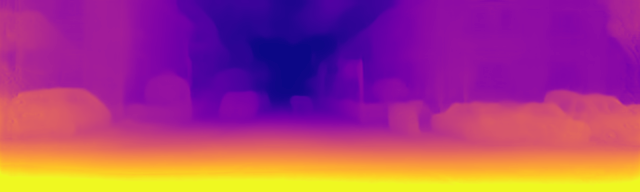}};
        \advance\vY by \vdY
        \node at (\vX,\vY) {\Pa{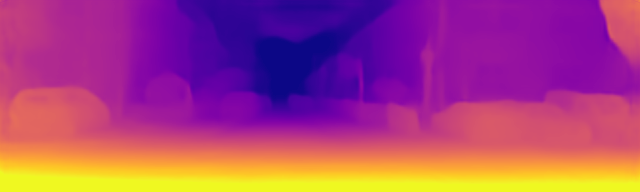}};

        \advance\vX by \vdX
        \vY = 0
        \node at (\vX,\vY) {\Pa{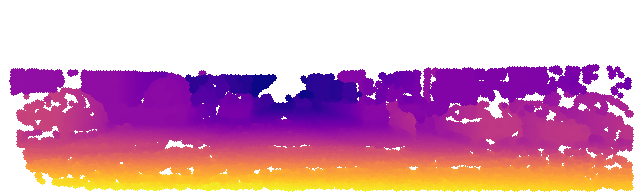}};

        \advance\vX by \vdXGT
        \advance\vY by \vdYGT
        \node[inner sep=0pt, font=\tiny] at (\vX,\vY) {GT};
        \advance\vX by -\vdXGT
        \advance\vY by -\vdYGT

        \advance\vY by \vdY
        \node at (\vX,\vY) {\Pa{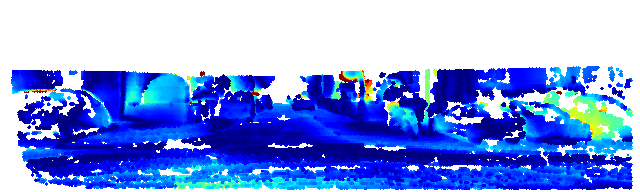}};

        \advance\vX by \vdXErr
        \advance\vY by \vdYErr
        \node[inner sep=0pt, font=\tiny] at (\vX,\vY) {Error map};
        \advance\vX by -\vdXErr
        \advance\vY by -\vdYErr

        \advance\vY by \vdY
        \node at (\vX,\vY) {\Pa{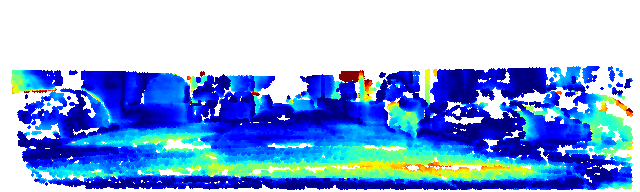}};
        \advance\vY by \vdY
        \node at (\vX,\vY) {\Pa{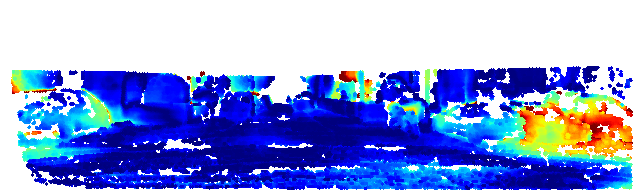}};
        \advance\vY by \vdY
        \node at (\vX,\vY) {\Pa{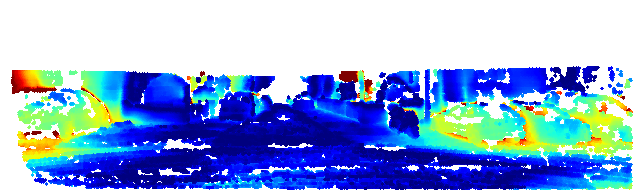}};

    \end{tikzpicture}
    \caption{Qualitative comparison on KITTI. In error maps, the larger depth errors are represented in red, smaller ones are in blue. The model trained with our FUMET predicts more accurate depth maps compared to the weakly-supervised methods.}
    \label{fig:QualityKITTI}
\end{figure*}

\subsubsection{KITTI.}
Our FUMET is independent of network architectures so we experiment on several models with FUMET as well as Monodepth2: Lite-Mono~\cite{Lite-Mono} with ResNet18, HR-Depth~\cite{HR-Depth} with ResNet18, and VADepth~\cite{VADepth}.
We train these models on the KITTI raw dataset~\cite{KITTI}, which contains 39,810 images for training and 4,424 for validation, and evaluate it with the KITTI Eigen test split~\cite{Eigen} (697 images). For training and evaluation, all images are resized to 640$\times$192.

We show the qualitative results in \cref{fig:QualityKITTI}. The quantitative results shown in \cref{tab:KITTIQuantity} demonstrate that FUMET enables the various architectures to learn metric scale properly.
Even when employing the simplest network architecture, \ie, Monodepth2, it outperforms weakly-supervised methods that rely on ground-truth scale labels for training which is not necessary for our method. Similar to VADepth~\cite{VADepth}, FUMET uses the camera height invariance as supervision though VADepth requires the ground-truth camera height and employs a complex network architecture. FUMET with Monodepth2 (row 6 in \cref{tab:KITTIQuantity}) achieves higher accuracy. This indicates that measuring camera height precisely is difficult and using unreliable ground-truth camera height causes low accuracy in the scale of estimated depth maps.

In the results showing the accuracy of the estimated depth map scaled with ground-truth scale factors via median scaling~\cite{SfMLearner}, it is evident that every model trained with FUMET attains superior results to the one trained without FUMET. 
This shows that our training framework not only enables the models to learn metric scale but also enhances the geometric accuracy.
This improvement owes to the camera height loss $\loss_\cam$ and $\loss_\aux$, which makes the model respect road regions and object regions, respectively.

FUMET successfully enables a model to learn metric scale, even though the training dataset of KITTI contains many frames with no or few detected cars. We detail this in the supplementary material.

\subsubsection{Cityscapes.}
To evaluate the ability of FUMET to leverage any dataset for training, we train and evaluate it on the Cityscapes dataset~\cite{Cityscapes}. We follow~\cite{SfMLearner, LEGO, GeoNet, ManyDepth} and collect 69,731 images for training and evaluate on the 1,525 test images. In this dataset, there are multiple sets of camera intrinsics. Since this strongly affects the metric depth estimation~\cite{Metric3D}, we first align the focal lengths of all images to 587.5 by resizing them, and crop them to 512$\times$192 resolution. For evaluation, we follow~\cite{Struct2Depth, ManyDepth} and further crop the estimated image into 416$\times$128.
In order to align the number of iterations on KITTI, we train for 29 epochs with the same learning rate decreased by half every 8 epochs, and $\tau_\mathrm{mid}$ is set to 12.

\cref{tab:Cityscapes} shows the quantitative results and FUMET outperforms weakly-supervised methods by wide margins.
These methods assume that highly accurate sensor data is available and fail to train scale-aware MDE models on datasets consisting of unreliable sensor data such as Cityscapes. On the other hand, FUMET does not require any sensor measurement other than RGB video, so it can robustly leverage any driving videos for training.

\subsubsection{Mixed datasets.}
To demonstrate that FUMET can train a model on a mixed set of datasets to attain strong generalizability, we train with 235,341 images collected from the Argoverse2~\cite{Argoverse2}, Lyft~\cite{Lyft}, A2D2~\cite{A2D2}, and DDAD~\cite{PackNet} dataset and evaluate it on the KITTI Eigen test split~\cite{Eigen}. Both images for training and evaluation are resized and cropped to 832$\times$512 resolution. 
The training images are the sets of video sequences and each sequence is captured with its own unique camera setting.
We individually optimize each camera height for each sequence during training.
We detail the dataset composition and training settings in the supplementary material.

As shown in \cref{tab:Multi}, the model trained with FUMET on the large-scale mixed dataset achieves comparable accuracy to the one trained on KITTI despite the domain gap and attains high generalizability. The model trained on both the mixed dataset and KITTI outperforms the model trained only on KITTI. This suggests that training on various images enhances accuracy which is the strength of fully self-supervised learning methods. 

We also show that FUMET can leverage in-the-wild videos for training by using camera intrinsics recovered with SLAM, \eg, COLMAP~\cite{COLMAP} in the supplementary material.

\begin{table}[t!]
  \begin{minipage}[t]{.48\textwidth}
    \caption{Quantitative results on Cityscapes~\cite{Cityscapes}. We train all the models from scratch with the same supervisions as the Supervision column in \cref{tab:KITTIQuantity}. FUMET significantly outperforms weakly-supervised methods.}
    \label{tab:Cityscapes}
    \centering
    {
    \scriptsize
    \setlength{\tabcolsep}{0.3em}
    \begin{tabular}{@{}c|ccc|c@{}}
    \toprule
    
    Method                 &
    \!AbsRel\!\!                          &
    \!\!SqRel\!\!                          &
    \!\!RMSE\!                           &
    \!$\delta\!\!<\!\!1.25$\!  \\
    
    \midrule
    G2S\!~\cite{GPS}                             & \!4.156 & \!\!276.16 & \!\!58.89 & 0.046 \\
    PackNet-SfM\!~\cite{PackNet}\!               & \!0.504 & \!\!6.639 & \!\!14.90 & 0.029 \\
    VADepth\!~\cite{VADepth}                   & \!0.363 & \!\!7.115 & \!\!11.95 & 0.295 \\

    \midrule

    \textbf{FUMET}               & \!\textbf{0.125} & \!\!\textbf{1.288} & \!\!\textbf{6.359} & \textbf{0.858} \\
    \bottomrule
    \end{tabular}
    }
  \end{minipage}
  \hfill
  \begin{minipage}[t]{.48\textwidth}
        \caption{FUMET can be trained on mixed datasets (Argoverse2, Lyft, A2D2, and DDAD) with various camera settings, and attains high generalizability. The results are evaluated on the KITTI Eigen split~\cite{Eigen} (832$\times$512 resolution).}
    \label{tab:Multi}
    \centering
    {
    \scriptsize
    \setlength{\tabcolsep}{0.3em}
    \begin{tabular}{@{}c|ccc|c@{}}
    \toprule
    Training dataset                   &
    \!AbsRel\!\!                         &
    \!\!SqRel\!\!                          &
    \!\!RMSE\!                           &
    \!$\delta\!<\!1.25$\!  \\
    \midrule
    KITTI   & 0.103 & 0.675 & 4.708 & 0.903 \\
    Mixed       & 0.113 & 0.916 & 5.009 & 0.883 \\
    Mixed\,$+$\,KITTI & \textbf{0.082} & \textbf{0.611} & \textbf{4.307} & \textbf{0.923} \\
    \bottomrule
    \end{tabular}
    }
  \end{minipage}
\end{table}

\subsubsection{Ablation Studies on KITTI.}
To study the effect of the proposed components in FUMET, we conduct ablation studies on the training losses, balancing the loss weights, camera height optimization, and outlier filtering. The results are shown in \cref{tab:AblationLossBalOpt}. Thanks to the accuracy of LSP and the robustness of Silhouette Projector, the camera height loss makes a greater contribution to the depth accuracy than the auxiliary rough geometric loss. By introducing the camera height optimization, the model can learn the scale with more accurate and consistent supervision than leveraging the size prior independently for each frame (rows 1, 2, and 6 in \cref{tab:AblationLossBalOpt}), which leads to improved accuracy. Though balancing the loss weights (\cref{eq:BalanceWeight}) with either $\loss_\aux$ or $\loss_\cam$ degrades the results due to the insufficient strength of the supervisory signals for metric scale, it contributes to improving accuracy when both $\loss_\aux$ and $\loss_\cam$ are employed.

Our proposed method can also be used with a pre-trained MDE model in an offline manner. We first train the original Monodepth2~\cite{Monodepth2} without FUMET, then predict depth maps of training images. With Silhouette Projector and LSP, we obtain per-frame \emph{scaled} camera height and optimize it across frames with the median operation. We fix this value and use it as the camera height supervision for an untrained Monodepth2 on FUMET. 
We examine the difference between the proposed camera height optimized online and the fixed one derived offline when we use them as the pseudo-supervision.
Note that we also fix $\lambda_\cam$ because the pseudo-label can be considered sufficiently accurate from the early stage of training. We also run another experiment in which we fix camera height supervision derived offline for the first 20 epochs, which is the same number of training epochs as the original Monodepth2, then unfix the camera height pseudo-label to fine-tune it via online optimization. As shown in \cref{tab:FixCamH}, the model supervised with the fixed camera height computed offline achieves slightly better results than the one supervised with the camera height optimized online. This result indicates that our strategy of deriving \emph{scaled} camera height is highly accurate and fixing it stabilizes the training. The model trained with the fine-tuned camera height achieves the highest accuracy, which shows that our FUMET can attain better accuracy with additional pre-training costs.

\begin{table}[t!]
  \begin{minipage}[t]{.51\textwidth}
    \caption{Each loss contributes to learning metric scale and the others enhance the depth accuracy. Bal: balancing the loss weights (\cref{eq:BalanceWeight}). Opt: camera height optimization. Flt: outlier filtering.}
    \label{tab:AblationLossBalOpt}
    \centering
    {
    \scriptsize
    \setlength{\tabcolsep}{0.3em}
    \begin{tabular}{@{}ccccc|ccc|c@{}}
    \toprule
    $\loss_\aux$\!\!                   &
    \!\!$\loss_\cam$\!\!                  &
    \!\!Bal\!\!                   &
    \!Opt\!\!                 &
    \!Flt\!                 &

    \!AbsRel\!\!                         &
    \!\!SqRel\!\!                          &
    \!\!RMSE\!                           &
    \!$\delta\!\!<\!\!1.25$  \\
    \midrule
    \checkmark & -          &  -         & -          & \checkmark & 0.125 & 0.887 & 5.055 & 0.834  \\
    \checkmark & -          & \checkmark & -          & \checkmark & 0.149 & 0.950 & 5.219 & 0.785  \\
    -          & \checkmark &  -         & \checkmark & \checkmark & 0.116 & 0.875 & 4.874 & \underline{0.863} \\
    -          & \checkmark & \checkmark & \checkmark & \checkmark & 0.211 & 1.303 & 6.143 & 0.585 \\
    \checkmark & \checkmark & -          & \checkmark & \checkmark & 0.122 & 0.880 & 5.017 & 0.849 \\
    \checkmark & \checkmark & \checkmark & -          & \checkmark & \underline{0.115} & \underline{0.822} & \underline{4.811} & 0.858 \\
    \checkmark & \checkmark & \checkmark & \checkmark & -          & 0.116 & 0.823 & 4.901 & 0.860 \\
    \checkmark & \checkmark & \checkmark & \checkmark & \checkmark & \textbf{0.108} & \textbf{0.785} &\!\! \textbf{4.736} \!& \textbf{0.871} \\

    \bottomrule
    \end{tabular}
    }
  \end{minipage}
  \hfill
  \begin{minipage}[t]{.46\textwidth}
    \caption{
    Training with a fine-tuned pseudo camera height achieves the highest accuracy. Online: optimizing the camera height while training. Offline: training with a fixed one computed from depth maps predicted with pre-trained Monodepth2. Fine-tune: training with the fixed one computed offline for 20 epochs, then optimizing online.
    }
    \label{tab:FixCamH}
    \centering
    {
    \scriptsize
    \setlength{\tabcolsep}{0.3em}
    \begin{tabular}{@{}c|ccc|c@{}}
    \toprule
    Deriv. method\!                   &
    \!AbsRel\!\!                         &
    \!\!SqRel\!\!                          &
    \!\!RMSE\!\!                           &
    \!$\delta\!<\!1.25$\!  \\
    \midrule
    Online   & 0.108 & 0.808 & 4.740 & 0.870 \\
    Offline       & \underline{0.106} & \underline{0.790} & \underline{4.719} & \underline{0.873} \\
    Fine-tune & \textbf{0.105} & \textbf{0.782} & \textbf{4.682} & \textbf{0.876} \\
    \bottomrule
    \end{tabular}
    }
  \end{minipage}
\end{table}


\subsection{LSP}

We provide implementation details of LSP in the supplementary material.

\subsubsection{Evaluation on KITTI and Cityscapes.}
We evaluate the generalizability of our LSP with the dimensions of the annotated 3D bounding box on both the KITTI~\cite{KITTI} and the Cityscapes~\cite{Cityscapes} datasets in the \emph{Car} class. We compare the results of LSP to a fixed car height prior (1.59m, following~\cite{PuttingObjects}) and MonoCInIS~\cite{MonoCInIS}, a 3D object detection model that outputs object dimensions from an image and an instance mask. We train MonoCInIS from scratch with the KITTI dataset. As an evaluation metric, we compute the absolute relative error for each dimension. When we estimate car dimensions with LSP and MonoCInIS, we use a ground-truth instance mask for making a car image mask and as input, respectively. For all the experiments in LSP, input vehicle masks are cropped and resized to 300$\times$300 resolution. On the KITTI dataset, we extract 3,769 samples for evaluation~\cite{KITTI3dSplit}. On the Cityscapes dataset, we evaluate on the validation split, which has 500 samples. On the KITTI dataset, each object is separated into 3 levels based on detection difficulty: Easy, Moderate, and Hard. Considering the difference in focal lengths between KITTI and Cityscapes, we also separate objects on Cityscapes into the same levels as on KITTI. 

As shown in \cref{tab:LearnedSizePrior}, LSP outperforms the other two methods in most cases. Particularly, the accuracy of MonoCInIS drops on Cityscapes since it overfits to the training dataset, \ie, KITTI and cannot address the domain gap. In contrast, thanks to the large training data and various augmentations, our LSP obtains the highest generalizability and robustly estimates dimensions on any dataset.
Learning not only car height but also width and length also improves the prediction accuracy as this facilitates estimation of the overall car sizes.

\begin{table*}[t!]
    \caption{Generalizability on KITTI and Cityscapes. We evaluate LSP on Absolute Relative Error for each dimension (height/width/length), comparing with the fixed car height prior (1.59 m, following~\cite{PuttingObjects}) and MonoCInIS~\cite{MonoCInIS} trained on KITTI. LSP robustly surpasses other methods in most cases and can predict accurately regardless of the datasets.}
    \label{tab:LearnedSizePrior}
    \centering
    \tiny
    \setlength{\tabcolsep}{0.3em}
    \begin{tabular}{l|c|c@{/}c@{/}cc@{/}c@{/}cc@{/}c@{/}cc@{/}c@{/}c}
    \toprule
    & Method                             & \multicolumn{3}{c}{Easy} & \multicolumn{3}{c}{Moderate}       & \multicolumn{3}{c}{Hard}           & \multicolumn{3}{c}{All} \\
    \midrule

    \parbox[t]{2mm}{\multirow{4}{*}{\rotatebox[origin=c]{90}{\!\!KITTI}}}

    & Fixed Height~\cite{PuttingObjects} & 0.065                    & -              & -              & \underline{0.064}          & -              & -              & \textbf{0.065} & -              & -              & \underline{0.065}             & -               & -              \\
    & MonoCInIS~\cite{MonoCInIS}         & 0.069                    & \textbf{0.065} & \textbf{0.093} & 0.112          & \textbf{0.106} & \textbf{0.133} & 0.268          & 0.263          & 0.269          & 0.159             & 0.155           & 0.177          \\
    \cmidrule{2-14}
    & LSP (only height)   &  \underline{0.057} & - & - & 0.089 & - & - & 0.106 & - & - & 0.087 & - & - \\
    & \textbf{LSP}        & \textbf{0.046}           & 0.136          & 0.130          & \textbf{0.060} & 0.144          & 0.139          & \underline{0.073}          & \textbf{0.155} & \textbf{0.150} & \textbf{0.062}    & \textbf{0.147}  & \textbf{0.145} \\
    \bottomrule
    \toprule
    \parbox[t]{2mm}{\multirow{4}{*}{\rotatebox[origin=c]{90}{Cityscapes\!}}}
    & Fixed Height~\cite{PuttingObjects} & 0.093                    & -              & -              & 0.093          & -              & -              & \underline{0.076}          & -              & -              & \underline{0.092}             & -               & -              \\
    & MonoCInIS~\cite{MonoCInIS}         & 0.578                    & 0.599          & 0.584          & 0.759          & 0.766          & 0.767          & 0.843          & 0.852          & 0.852          & 0.851             & 0.859           & 0.856          \\
    \cmidrule{2-14}
    & LSP (only height)   & \underline{0.063} & - & - & \underline{0.075} & - & - & 0.103 & - & - & 0.096 & - & - \\
    & \textbf{LSP}        & \textbf{0.056}           & \textbf{0.069} & \textbf{0.067} & \textbf{0.068} & \textbf{0.068} & \textbf{0.078} & \textbf{0.069} & \textbf{0.059} & \textbf{0.075} & \textbf{0.080}    & \textbf{0.064}  & \textbf{0.077} \\

    \bottomrule
    \end{tabular}
\end{table*}

\section{Conclusion}
In this work, we proposed FUMET, a novel training framework for any MDE model to learn metric scale.
FUMET aggregates the scale information obtained from a learned car size prior into the camera height and uses it as pseudo-supervision formulated as optimization across frames and epochs by leveraging its invariance. We showed that FUMET enables monocular depth estimators to embody metric scale regardless of the underlying network architectures and achieves state-of-the-art absolute-scale depth estimation accuracy on both the KITTI and the Cityscapes datasets, and demonstrated that it can leverage any dataset for training. We also proposed LSP, which can be easily trained on a large-scale dataset without manual annotations and robustly estimates vehicle dimensions from their appearances. We showed that the learned size prior is superior to a fixed one and outperforms a similar 3D object detection model. FUMET can be leveraged as a drop-in training scheme for any monocular depth estimation method. We hope it will be a versatile tool for road scene understanding research. 

\section*{Acknowledgements}
This work was in part supported by
JSPS 
20H05951 and 
21H04893, 
JST JPMJCR20G7 
and JPMJAP2305, 
and RIKEN GRP.

\bibliographystyle{splncs04}
\bibliography{main}

\clearpage

\appendix

\section{Details of Training on the Mixed Datasets}
\label{sec:Multi-dataset}

\subsection{Dataset Preparation}
We use the images from the front camera in Argoverse2~\cite{Argoverse2}, Lyft~\cite{Lyft}, A2D2~\cite{A2D2}, and DDAD~\cite{PackNet} for training. As mentioned in \cref{sec:FUMETResults} since the focal lengths are essential for metric MDE, we resize the image and align the focal lengths to 1000 px. In order to facilitate the training process, we crop the images to 832$\times$512 resolution. These resizing and cropping are also done for evaluation on the KITTI Eigen split~\cite{Eigen}.

To make the training easier and more efficient, we eliminate frames where the camera is stationary since training with these frames greatly degrades the inference performance~\cite{Monodepth2}. We achieve this elimination by looking at pairs of successive frames and simply checking the number of pixels with large differences in normalized intensities.
Images with a small number of pixels with large intensity changes are deemed stationary and discarded from the training dataset.
Since Argoverse2 is captured at a higher frame rate than the other datasets, we treat every other frame as successive frames.

For optimization during training, FUMET requires a sufficient number of images captured with the same camera height. We select such sequences with at least 2,500 frames. At the end of each training epoch, we optimize each camera height individually. In total, we use 235,341 images for training.

\subsection{Training Settings}
\label{sec:MixedDatasetSettings}
We train Monodepth2~\cite{Monodepth2} with ResNet50 on FUMET with the same batch size and initial learning rate as used during training on the KITTI dataset (\cref{sec:FUMETResults}). Considering the gap of training samples between KITTI and the mixed dataset, we set $\tau_\mathrm{mid}=4$ and train for 8 epochs while reducing the learning rate by half every 3 epochs. For training on the mixed dataset and KITTI, we train for 8 epochs with $\tau_\mathrm{mid}=3$ and reduce the learning rate by half every 2 epochs.


\section{Camera Intrinsics Recovered with COLMAP}
Similar to previous self-supervised MDE methods, FUMET requires camera intrinsics to compute the photometric error in training. In the following section, we demonstrate that FUMET successfully trains a model with intrinsics recovered with COLMAP~\cite{COLMAP} on the KITTI dataset~\cite{KITTI}. In~\cref{sec:YouTube}, we also verify that FUMET can leverage in-the-wild videos for training by using COLMAP.

\subsection{Training on KITTI with Recovered Camera Intrinsics}

We run COLMAP with the pinhole camera assumptions on KITTI sequences containing 500 to 800 frames and use the median of the computed intrinsics for training. We train Monodepth2~\cite{Monodepth2} (ResNet50) on FUMET under the same training settings as described in \cref{sec:FUMETResults}. The results on KITTI (\cref{tab:COLMAP}) show that thanks to the accuracy of COLMAP, the model trained with the recovered intrinsics achieves comparable accuracy to that trained with the ground-truth values.

\begin{table}[t!]
    \caption{
        Comparison of FUMET trained with ground-truth camera intrinsics versus the one trained with recovered values with COLMAP. Each model is trained and evaluated on KITTI~\cite{KITTI,Eigen} (640$\times$192 resolution).
        FUMET can train a model with recovered camera intrinsics, achieving accuracy comparable to that trained with ground-truth ones.
    }
    \label{tab:COLMAP}
    \centering
    {
    \footnotesize
    \setlength{\tabcolsep}{0.3em}
    \begin{tabular}{@{}c|cccc|ccc@{}}
    \toprule

                              & \multicolumn{4}{c|}{Error ($\downarrow$) }                            & \multicolumn{3}{c}{Accuracy ($\uparrow$)} \\
    Intrinsics                & AbsRel & SqRel & RMSE & RMSE$_{\mathrm{log}}$ & $\delta\!<\!1.25$ & $\delta\!<\!1.25^2$ & $\delta\!<\!1.25^3$ \\
    \midrule

    GT     & \textbf{0.108} & 0.785 & 4.736 & \textbf{0.195} & \textbf{0.871} & \textbf{0.958} & \textbf{0.981} \\
    COLMAP & 0.110 & \textbf{0.774} & \textbf{4.721} & 0.197 & 0.867 & \textbf{0.958} & \textbf{0.981} \\

    \bottomrule
    \end{tabular}
    }
\end{table}

\subsection{Training with In-the-wild Videos}
\label{sec:YouTube}

We download two YouTube videos \cite{Raleigh,Uruguay} (73,100 frames) and run COLMAP to recover the intrinsics. Since these videos are too long to use all the frames for COLMAP, we manually extract consecutive 300 frames from uncrowded scenes and run COLMAP individually with the pinhole camera assumptions.
We train Monodepth2 with FUMET on the dataset composed of the mixed dataset (Argoverse2~\cite{Argoverse2}, Lyft~\cite{Lyft}, A2D2~\cite{A2D2}, and DDAD~\cite{PackNet}), KITTI and the YouTube videos. The batch size and the initial learning rate are the same as those used in \cref{sec:FUMETResults}.
For the same reason as described in~\cref{sec:MixedDatasetSettings}, we train for 6 epochs with $\tau_\mathrm{mid}=2$, reducing the learning rate by half every 2 epochs.

The results on KITTI shown in \cref{tab:YouTube} demonstrate that adding in-the-wild videos to the training dataset increases the dataset diversity and improves the depth accuracy. This suggests the significance of being able to train without sensor measurements.

\begin{table}[t!]
    \caption{
        FUMET can leverage YouTube videos for training and lead to slightly improved accuracy. The results are evaluated on KITTI~\cite{Eigen} (832$\times$512 resolution). In the Dataset column, M, K, and Y represent the mixed dataset, KITTI, and YouTube, respectively.
    }
    \label{tab:YouTube}
    \centering
    {
    \footnotesize
    \setlength{\tabcolsep}{0.3em}
    \begin{tabular}{@{}c|cccc|ccc@{}}
    \toprule

                              &  \multicolumn{4}{c|}{Error ($\downarrow$) }                           & \multicolumn{3}{c}{Accuracy ($\uparrow$)} \\
    Dataset                   &  AbsRel & SqRel & RMSE & RMSE$_{\mathrm{log}}$ & $\delta\!<\!1.25$ & $\delta\!<\!1.25^2$ & $\delta\!<\!1.25^3$ \\
    \midrule

    M \!$+$\! K                              & \textbf{0.082} & 0.611 & 4.307 & 0.149 & \textbf{0.923} & 0.976 & \textbf{0.989}  \\
    M \!$+$\! K \!$+$\! Y          & 0.084 & \textbf{0.608} & \textbf{4.264} & \textbf{0.148} & \textbf{0.923} & \textbf{0.977} & \textbf{0.989} \\

    \bottomrule
    \end{tabular}
    }
\end{table}


\section{Assumption of Vehicle Presence}
\begin{figure}
  \centering
  \includegraphics[width=0.6\linewidth]{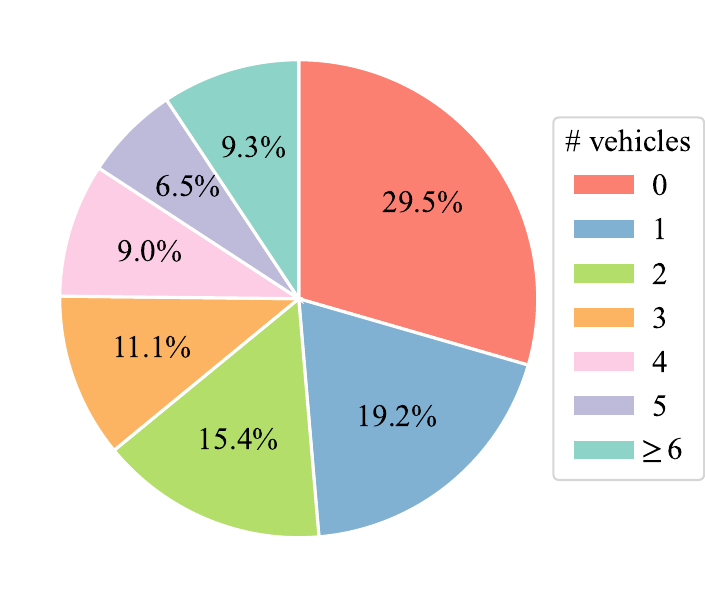}
  \caption{
    The proportion of frames with the number of observed cars in the training dataset of KITTI~\cite{KITTI}. There is no or only one observed car in nearly half of the frames in the dataset,
  }
  \label{fig:KITTIVehiclePie}
\end{figure}

In contrast to inference time, FUMET leverages observed vehicles and creates pseudo scale supervision for training. However, it does not create pseudo supervision for each frame independently but instead aggregates the scale information into the camera height and optimizes it across the sequence. This allows FUMET to create scale supervision from frames observing vehicles and provide scale supervisory signals even for frames without observed vehicles. In fact, as shown in \cref{fig:KITTIVehiclePie}, although the training dataset of KITTI contains many frames where vehicles are not observed or are scarce, FUMET can train with high depth accuracy.

\section{Comparison to a Supervised Method}
Our key contribution is a self-supervision cue for metric depth reconstruction. As such, direct comparison with supervised methods (\eg, ZoeDepth~\cite{ZoeDepth}) would be unfair. That said, we can compare with supervised methods trained on our mixed dataset using LiDAR depth. We use ZoeDepth as a representative method for this comparison. We select the architecture most similar to our default model (ResNet50) from its official implementation in terms of the number of parameters and the architecture type, \ie, ConvNet, and skip pre-training performed in the original paper for fairness.

The results on KITTI (\cref{tab:ZoeDepth}) show that FUMET slightly outperforms ZoeDepth. This suggests that it can achieve accuracy comparable to supervised methods. This is likely because the richer self-supervision from the reconstruction loss (\cref{eq:RecLoss}) forces the model to recognize scene contexts effectively, more so than sparse LiDAR depth, and enables metric scale inference from them without overfitting.

\begin{table}[t!]
    \caption{
        Comparison with ZoeDepth~\cite{ZoeDepth} trained on the mixed dataset with LiDAR depth. The results are evaluated on KITTI~\cite{Eigen} (832$\times$512 resolution). FUMET slightly outperforms ZoeDepth in terms of depth accuracy even though FUMET does not use ground-truth depth maps.
    }
    \label{tab:ZoeDepth}
    \centering
    {
    \footnotesize
    \setlength{\tabcolsep}{0.3em}
    \begin{tabular}{@{}c|c|cccc|ccc@{}}
    \toprule

                              &             &  \multicolumn{4}{c|}{Error ($\downarrow$) }                           & \multicolumn{3}{c}{Accuracy ($\uparrow$)} \\
    Method                    & \!Supervision\! & \!AbsRel\!\! & \!SqRel\!\! & \!\!RMSE\!\! & \!RMSE$_{\mathrm{log}}$\! & \!$\delta\!\!<\!\!1.25$\!\! & \!$\delta\!\!<\!\!1.25^2$\!\! & \!$\delta\!\!<\!\!1.25^3$\! \\
    \midrule

    ZoeDepth\!~\cite{ZoeDepth}         & GT & 0.120 & \textbf{0.743} & 5.118 & 0.182 & 0.870 & \textbf{0.964} & \textbf{0.988} \\
    \textbf{FUMET}                            & -  & \textbf{0.113} & 0.916 & \textbf{5.009} & \textbf{0.181} & \textbf{0.883} & 0.961 & 0.983  \\

    \bottomrule
    \end{tabular}
    }
\end{table}


\section{Implementation Details of LSP}

We implement LSP with PyTorch~\cite{PyTorch}. We train it for 15 epochs with a batch size of 32 and warm-up the first 5 epochs. We use AdamW~\cite{AdamW} optimizer with a learning rate of $5.0\times 10^{-4}$ linearly decreased to $4.0\times 10^{-4}$ and a weight decay of 0.05. As a backbone network, we employ ConvNeXt V2~\cite{ConvNeXtV2} pre-trained with ImageNet22K~\cite{ImageNet} and concatenate two linear layers to it. As a training dataset, we use DVM-CAR~\cite{DVM-CAR}, which contains 1,451,784 vehicle mask images from 899 UK market car models and their dimensions. All images are resized to 300$\times$300 resolution. We use 80\% of the data for training and others for validation. We adopt multiple data augmentations mentioned in \cref{sec:LearnedSizePrior} to make the model robust to the domain gap. We use L1 loss for each dimension as the objective function.

\section{Ablation Studies of LSP}

\begin{figure*}
  \centering
  \includegraphics[width=\linewidth]{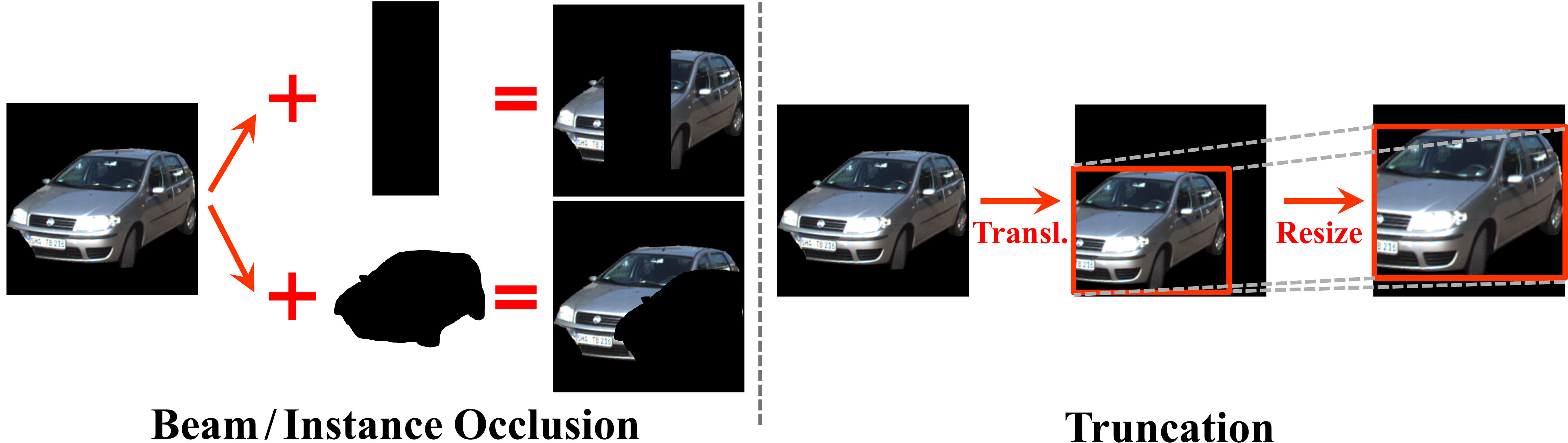}
  \caption{Visualization of beam/instance occlusion and truncation augmentations for training LSP.}
  \label{fig:Augmentations}
\end{figure*}

As specified in \cref{sec:LearnedSizePrior} in the main paper, during training of LSP, we perform beam/instance occlusion and truncation augmentations illustrated in ~\cref{fig:Augmentations} along with common color jittering and blur. The beam occlusion simulates objects occluded by poles including traffic lights and trees. Their segmentation mask is divided into more than two regions. The instance occlusion augmentation is for objects occluded by the other things including vehicles and people. Truncation augmentation is for objects that are partially cropped out of images. In ~\cref{tab:Augmentations}, we ablate the effectiveness of these augmentations. The results show that each augmentation contributes to improving the prediction accuracy both on the KITTI and the Cityscapes datasets, \ie, they enhance the robustness to any input image.

\begin{table*}[t!]
    \caption{
    Ablation of augmentations for LSP on KITTI~\cite{KITTI3dSplit} and Cityscapes~\cite{Cityscapes}. We evaluate each predicted dimension (height/width/length) on the absolute relative error. Both beam/instance occlusion and truncation augmentation improve the prediction accuracy. 
    }
    \label{tab:Augmentations}
    \centering
    \footnotesize
    \setlength{\tabcolsep}{0.3em}
    \begin{tabular}{ccc|c@{/}c@{/}c|c@{/}c@{/}c}
    \toprule
    beam occ. & inst. occ. & trunc. & \multicolumn{3}{c|}{KITTI} & \multicolumn{3}{c}{Cityscapes} \\
    \midrule
    
               &            &            & 0.081 & 0.159 & 0.174 & 0.088 & 0.065 & 0.082 \\ 
               & \checkmark & \checkmark & \underline{0.071} & 0.162 & 0.179            &   \textbf{0.080} & 0.069 & 0.080 \\
    \checkmark &            & \checkmark & 0.074 & 0.150 & 0.172                        &   \underline{0.083} & 0.065 & \underline{0.078}      \\
    \checkmark & \checkmark &            & 0.079 & \textbf{0.136} & \underline{0.165}   &   0.084 & \textbf{0.057} & \textbf{0.077}  \\
    \midrule
    \checkmark & \checkmark & \checkmark & \textbf{0.062} & \underline{0.147} & \textbf{0.145}      &   \textbf{0.080} & \underline{0.064} & \textbf{0.077}   \\
    \bottomrule
    \end{tabular}
\end{table*}

\section{Additional Quantitative Results on KITTI}
In the main paper, we trained the multiple networks with FUMET on the KITTI dataset~\cite{KITTI} and evaluated them on the KITTI Eigen split~\cite{Eigen} which is the most widely used, comparing the weakly-supervised methods. We also evaluate the same models with the improved ground-truth depth maps for KITTI introduced in~\cite{ImprovedGT}. These improved depth maps are composed of 652 out of 697 test frames in the Eigen split. The results shown in \cref{tab:KITTIQuantityImproved} demonstrate that the models trained with FUMET still surpass weakly-supervised methods, both with and without the median scaling, as well as when evaluating on the KITTI Eigen split.
It should be emphasized that VADepth~\cite{VADepth} trained with FUMET (row 9 and 18) outperforms the same network supervised with the measured camera height (row 4 and 13).
This proves that the pseudo camera height obtained via the camera height optimization is more precise than the measured one.

\begin{table*}[t!]
    \caption{Evaluation on KITTI improved ground truth~\cite{ImprovedGT} (640$\times$192 resolution). The models trained with FUMET outperform weakly-supervised methods in both results, with and without the median scaling~\cite{SfMLearner}.}
    \label{tab:KITTIQuantityImproved}
    \centering
    {
    \scriptsize
    \setlength{\tabcolsep}{0.3em}
    \begin{tabular}{@{}c|c|c|cccc|ccc@{}}
    \toprule

                              &              &              & \multicolumn{4}{c|}{Error ($\downarrow$) } & \multicolumn{3}{c}{Accuracy ($\uparrow$)} \\
    Method                    & \!Supervision\!  & \!Scaling\!          & \!AbsRel\!\!            & \!SqRel\!\!             & \!\!RMSE\!\!              & \!RMSE$_{\mathrm{log}}$\! & \!$\delta\!\!<\!\!1.25$\!\!     & \!$\delta\!\!<\!\!1.25^2$\!\!   & \!$\delta\!\!<\!\!1.25^3$\! \\
    \midrule
    G2S\!~\cite{GPS}                                       & GPS         & - & 0.111 & 0.791 & 4.523 & 0.168 & 0.869 & 0.967 & 0.989 \\ 
    PackNet-SfM\!~\cite{PackNet}                           & V           & - & 0.105 & 0.656 & 4.098 & 0.156 & 0.886 & 0.971 & 0.991 \\
    Wagstaff \etal\!~\cite{ScaleRecovery}                  & CamH        & - & 0.097 & 0.636 & 4.281 & 0.153 & 0.887 & 0.975 & 0.993 \\
    VADepth\!~\cite{VADepth}                               & CamH        & - & 0.091 & 0.555 & 3.871 & \underline{0.134} & \textbf{0.913} & \underline{0.983} & \underline{0.995} \\
    DynaDepth\!~\cite{IMU}                                 & \!IMU \!\!+\!\! V \!\!+\!\! G\! & - & 0.092 & \textbf{0.477} & \underline{3.756} & 0.135 & 0.906 & \underline{0.983} & \textbf{0.996} \\
    \cmidrule{1-10}
    \textbf{Ours}                                                 & -           & - & \textbf{0.089} & \underline{0.506} & 3.806 & \underline{0.134} & 0.908 & \textbf{0.984} & \textbf{0.996} \\
    Lite-Mono\!~\cite{Lite-Mono} \!\!$+$\!\! \textbf{Ours} & -           & - & \underline{0.090} & 0.526 & 3.767 & \underline{0.134} & \underline{0.910} & \textbf{0.984} & \underline{0.995} \\
    HR-Depth\!~\cite{HR-Depth} \!\!$+$\!\! \textbf{Ours}   & -           & - & 0.096 & 0.524 & 3.852 & 0.139 & 0.902 & 0.982 & \textbf{0.996} \\
    VADepth\!~\cite{VADepth} \!\!$+$\!\! \textbf{Ours}     & -           & - & 0.101 & 0.547 & \textbf{3.675} & \textbf{0.133} & \underline{0.910} & \textbf{0.984} & \textbf{0.996} \\

    \toprule
    \toprule

    G2S\!~\cite{GPS}                                       & GPS         & \checkmark & 0.088 & 0.544 & 3.968 & 0.137 & 0.913 & 0.981 & 0.995 \\
    PackNet-SfM\!~\cite{PackNet}                           & V           & \checkmark & 0.100 & 0.606 & 3.943 & 0.145 & 0.900 & 0.977 & 0.993 \\
    Wagstaff \etal\!~\cite{ScaleRecovery}                  & CamH        & \checkmark & 0.095 & 0.602 & 4.145 & 0.146 & 0.902 & 0.978 & 0.994 \\
    VADepth\!~\cite{VADepth}                               & CamH        & \checkmark & \underline{0.080} & 0.518 & 3.745 & \underline{0.123} & \underline{0.933} & \underline{0.986} & \underline{0.996} \\
    DynaDepth\!~\cite{IMU}                                 & \!IMU \!\!+\!\! V \!\!+\!\! G\! & \checkmark & 0.084 & \underline{0.448} & 3.761 & 0.130 & 0.917 & 0.984 & \underline{0.996} \\
    \cmidrule{1-10}
    \textbf{Ours}                                                 & -           & \checkmark & 0.081 & 0.455 & 3.675 & 0.125 & 0.927 & \underline{0.986} & \underline{0.996} \\
    Lite-Mono\!~\cite{Lite-Mono} \!\!$+$\!\! \textbf{Ours} & -           & \checkmark & 0.083 & 0.478 & \underline{3.632} & 0.126 & 0.926 & \underline{0.986} & \underline{0.996} \\
    HR-Depth\!~\cite{HR-Depth} \!\!$+$\!\! \textbf{Ours}   & -           & \checkmark & 0.084 & 0.461 & 3.719 & 0.129 & 0.922 & 0.985 & \underline{0.996} \\
    VADepth\!~\cite{VADepth} \!\!$+$\!\! \textbf{Ours}     & -           & \checkmark & \textbf{0.077} & \textbf{0.426} & \textbf{3.455} & \textbf{0.116} & \textbf{0.937} & \textbf{0.989} & \textbf{0.997} \\

    \bottomrule
    \end{tabular}
    }
\end{table*}


\section{Additional Qualitative Results on KITTI}
\begin{figure}[t]
    \centering
    \def\Pa#1{\includegraphics[width=0.235\linewidth]{#1}}
\begin{tikzpicture}[x=0.99mm,y=1mm]
    \newcount\vX
    \newcount\vY
    \newcount\vdX
    \newcount\vdY
    \newcount\vXi
    \newcount\vYi
    \newcount\vdXDyna
    \newcount\vdXGT
    \newcount\vdYGT
    \newcount\vdXErr
    \newcount\vdYErr
    \newcount\gap
    \vdX = 29
    \vdY = -11
    \vXi = -16
    \vYi = 58
    \vdXDyna = 1
    \vdXGT = 12
    \vdYGT = 3
    \vdXErr = 8
    \vdYErr = 3
    \gap = 1

    \vX = \vXi
    \vY = 0
    \node[inner sep=0pt, font=\tiny] at (\vX,\vY) {\rotatebox[origin=c]{90}{\parbox[t]{20mm}{\centering Input}}};
    \advance\vY by \vdY
    \node[inner sep=0pt, font=\tiny] at (\vX,\vY) {\rotatebox[origin=c]{90}{\parbox[t]{20mm}{\centering \textbf{FUMET}}}};
    \advance\vY by \vdY

    \advance \vX by -\vdXDyna
    \node[inner sep=0pt, font=\tiny] at (\vX,\vY) {\rotatebox[origin=c]{90}{\parbox[t]{20mm}{\centering Dyna\\Depth\!\!~\cite{IMU}}}};
    \advance \vX by \vdXDyna

    \advance\vY by \vdY
    \node[inner sep=0pt, font=\tiny] at (\vX,\vY) {\rotatebox[origin=c]{90}{\parbox[t]{20mm}{\centering G2S\!\!~\cite{GPS}}}};
    \advance\vY by \vdY
    \node[inner sep=0pt, font=\tiny] at (\vX,\vY) {\rotatebox[origin=c]{90}{\parbox[t]{20mm}{\centering VADepth\!\!~\cite{VADepth}}}};
    \advance\vY by \vdY

    \vX = 0
    \vY = 0
    \node at (\vX,\vY) {\Pa{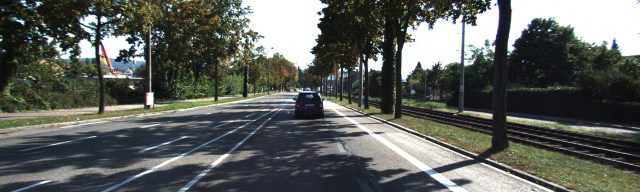}};
    \advance\vY by \vdY
    \node at (\vX,\vY) {\Pa{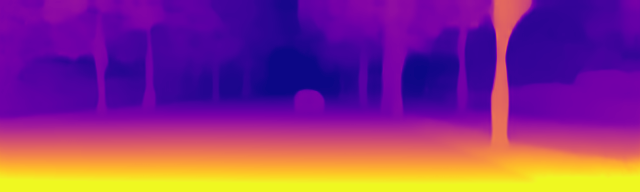}};
    \advance\vY by \vdY
    \node at (\vX,\vY) {\Pa{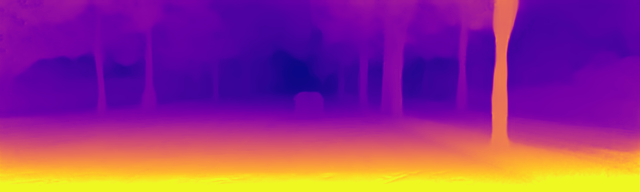}};
    \advance\vY by \vdY
    \node at (\vX,\vY) {\Pa{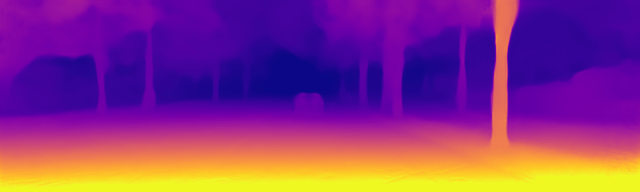}};
    \advance\vY by \vdY
    \node at (\vX,\vY) {\Pa{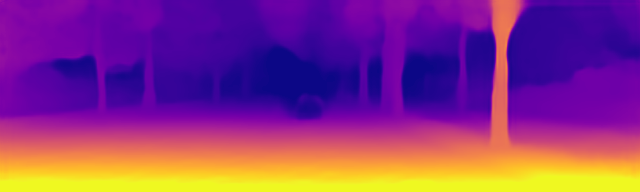}};

    \advance\vX by \vdX
    \vY = 0
    \node at (\vX,\vY) {\Pa{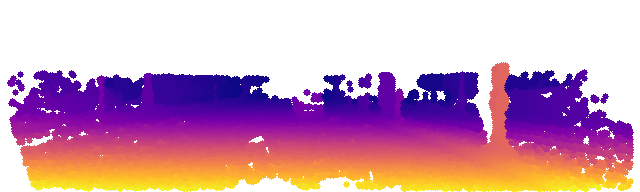}};

    \advance\vX by \vdXGT
    \advance\vY by \vdYGT
    \node[inner sep=0pt, font=\tiny] at (\vX,\vY) {GT};
    \advance\vX by -\vdXGT
    \advance\vY by -\vdYGT

    \advance\vY by \vdY
    \node at (\vX,\vY) {\Pa{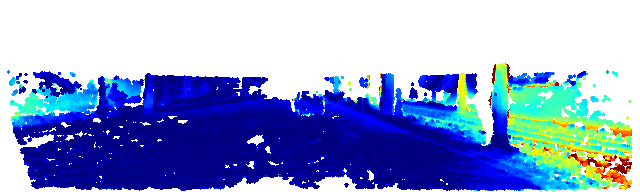}};

    \advance\vX by \vdXErr
    \advance\vY by \vdYErr
    \node[inner sep=0pt, font=\tiny] at (\vX,\vY) {Error map};
    \advance\vX by -\vdXErr
    \advance\vY by -\vdYErr

    \advance\vY by \vdY
    \node at (\vX,\vY) {\Pa{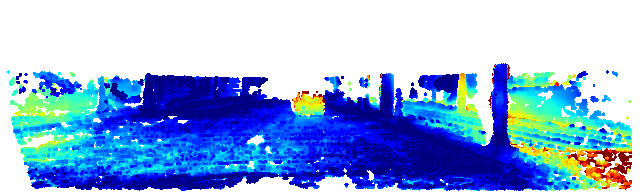}};
    \advance\vY by \vdY
    \node at (\vX,\vY) {\Pa{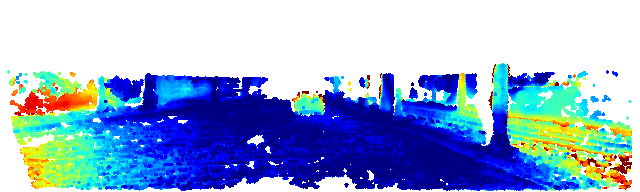}};
    \advance\vY by \vdY
    \node at (\vX,\vY) {\Pa{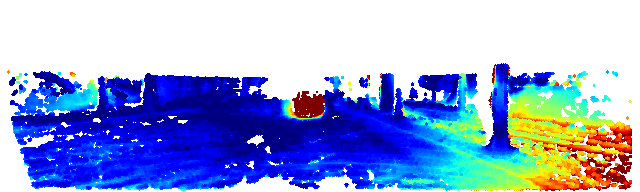}};

    \advance\vX by \vdX
    \advance\vX by \gap
    \vY = 0
    \node at (\vX,\vY) {\Pa{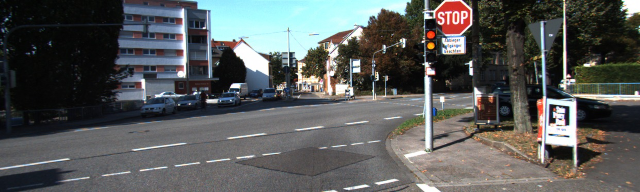}};
    \advance\vY by \vdY
    \node at (\vX,\vY) {\Pa{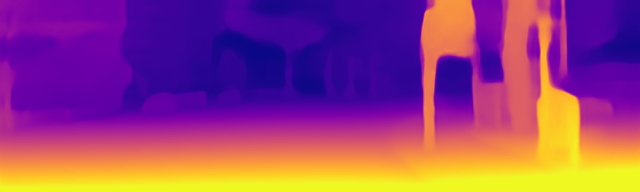}};
    \advance\vY by \vdY
    \node at (\vX,\vY) {\Pa{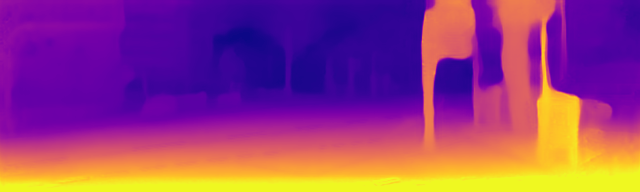}};
    \advance\vY by \vdY
    \node at (\vX,\vY) {\Pa{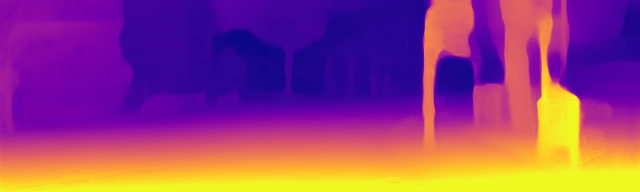}};
    \advance\vY by \vdY
    \node at (\vX,\vY) {\Pa{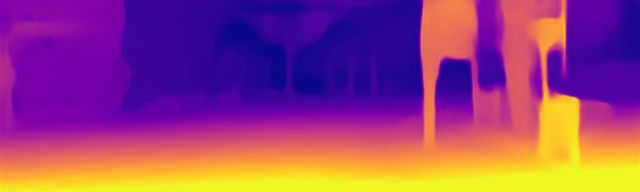}};

    \advance\vX by \vdX
    \vY = 0
    \node at (\vX,\vY) {\Pa{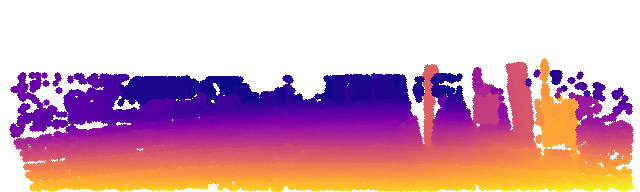}};

    \advance\vX by \vdXGT
    \advance\vY by \vdYGT
    \node[inner sep=0pt, font=\tiny] at (\vX,\vY) {GT};
    \advance\vX by -\vdXGT
    \advance\vY by -\vdYGT

    \advance\vY by \vdY
    \node at (\vX,\vY) {\Pa{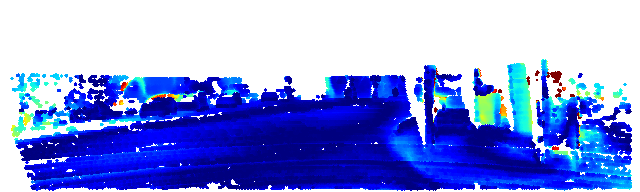}};

    \advance\vX by \vdXErr
    \advance\vY by \vdYErr
    \node[inner sep=0pt, font=\tiny] at (\vX,\vY) {Error map};
    \advance\vX by -\vdXErr
    \advance\vY by -\vdYErr

    \advance\vY by \vdY
    \node at (\vX,\vY) {\Pa{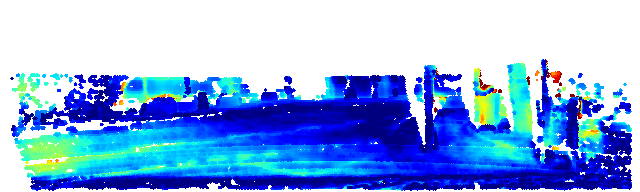}};
    \advance\vY by \vdY
    \node at (\vX,\vY) {\Pa{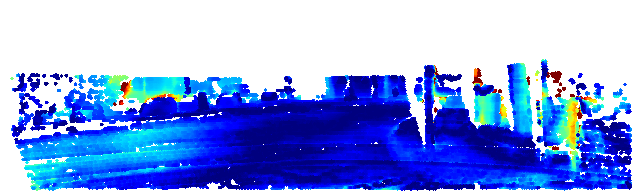}};
    \advance\vY by \vdY
    \node at (\vX,\vY) {\Pa{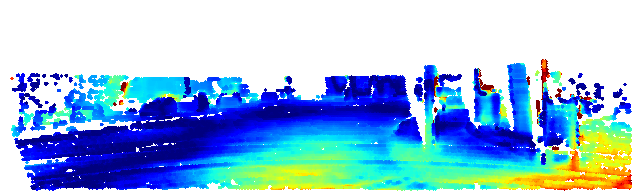}};

    \vX = \vXi
    \vY = \vYi
    \node[inner sep=0pt, font=\tiny] at (\vX,\vY) {\rotatebox[origin=c]{90}{\parbox[t]{20mm}{\centering Input}}};
    \advance\vY by \vdY
    \node[inner sep=0pt, font=\tiny] at (\vX,\vY) {\rotatebox[origin=c]{90}{\parbox[t]{20mm}{\centering \textbf{FUMET}}}};
    \advance\vY by \vdY

    \advance \vX by -\vdXDyna
    \node[inner sep=0pt, font=\tiny] at (\vX,\vY) {\rotatebox[origin=c]{90}{\parbox[t]{20mm}{\centering Dyna\\Depth\!\!~\cite{IMU}}}};
    \advance \vX by \vdXDyna

    \advance\vY by \vdY
    \node[inner sep=0pt, font=\tiny] at (\vX,\vY) {\rotatebox[origin=c]{90}{\parbox[t]{20mm}{\centering G2S\!\!~\cite{GPS}}}};
    \advance\vY by \vdY
    \node[inner sep=0pt, font=\tiny] at (\vX,\vY) {\rotatebox[origin=c]{90}{\parbox[t]{20mm}{\centering VADepth\!\!~\cite{VADepth}}}};
    \advance\vY by \vdY

    \vX = 0
    \vY = \vYi
    \node at (\vX,\vY) {\Pa{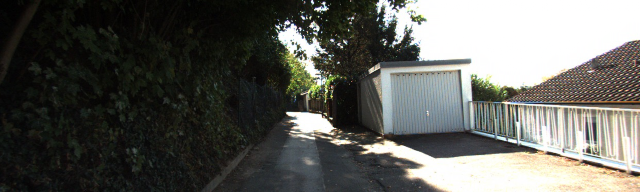}};
    \advance\vY by \vdY
    \node at (\vX,\vY) {\Pa{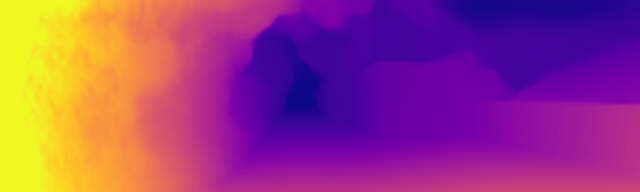}};
    \advance\vY by \vdY
    \node at (\vX,\vY) {\Pa{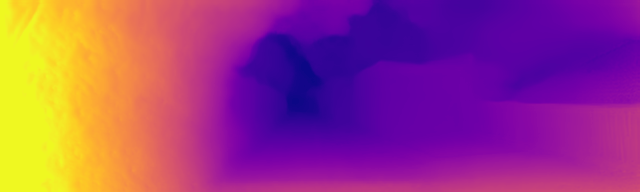}};
    \advance\vY by \vdY
    \node at (\vX,\vY) {\Pa{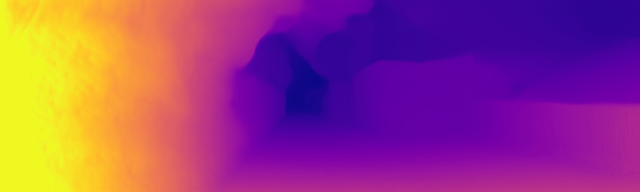}};
    \advance\vY by \vdY
    \node at (\vX,\vY) {\Pa{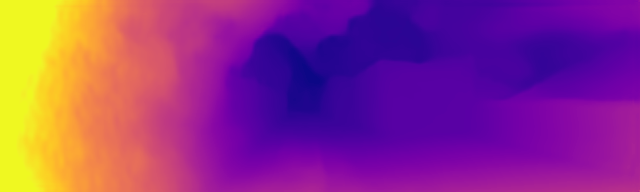}};

    \advance\vX by \vdX
    \vY = \vYi
    \node at (\vX,\vY) {\Pa{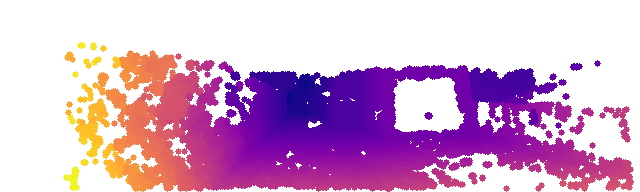}};

    \advance\vX by \vdXGT
    \advance\vY by \vdYGT
    \node[inner sep=0pt, font=\tiny] at (\vX,\vY) {GT};
    \advance\vX by -\vdXGT
    \advance\vY by -\vdYGT

    \advance\vY by \vdY
    \node at (\vX,\vY) {\Pa{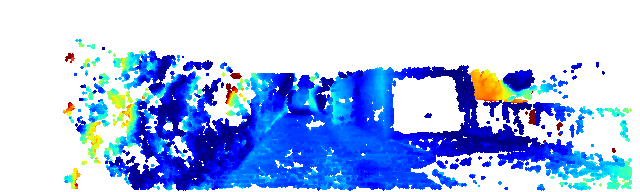}};

    \advance\vX by \vdXErr
    \advance\vY by \vdYErr
    \node[inner sep=0pt, font=\tiny] at (\vX,\vY) {Error map};
    \advance\vX by -\vdXErr
    \advance\vY by -\vdYErr

    \advance\vY by \vdY
    \node at (\vX,\vY) {\Pa{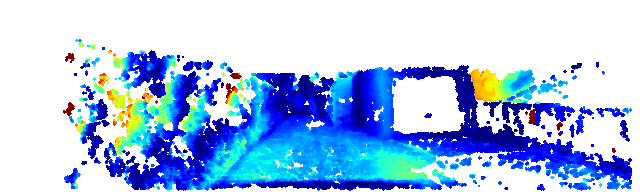}};
    \advance\vY by \vdY
    \node at (\vX,\vY) {\Pa{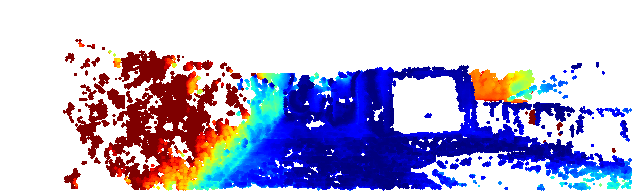}};
    \advance\vY by \vdY
    \node at (\vX,\vY) {\Pa{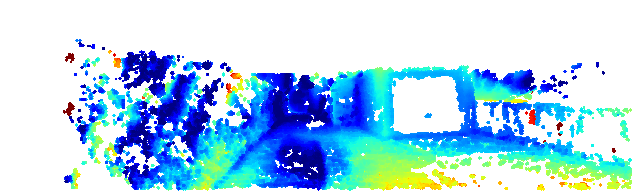}};

    \advance\vX by \vdX
    \advance\vX by \gap
    \vY = \vYi
    \node at (\vX,\vY) {\Pa{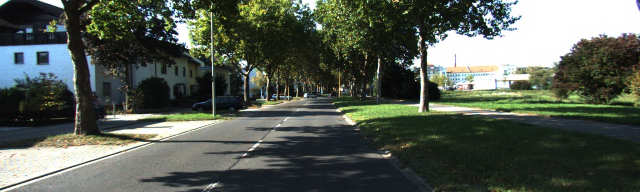}};
    \advance\vY by \vdY
    \node at (\vX,\vY) {\Pa{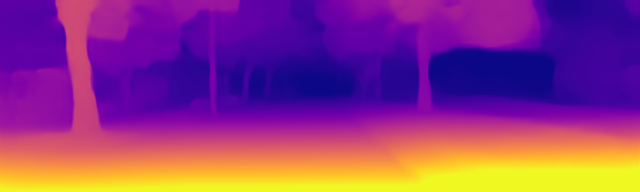}};
    \advance\vY by \vdY
    \node at (\vX,\vY) {\Pa{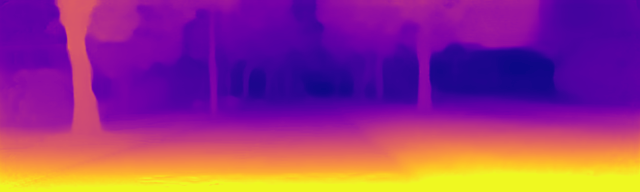}};
    \advance\vY by \vdY
    \node at (\vX,\vY) {\Pa{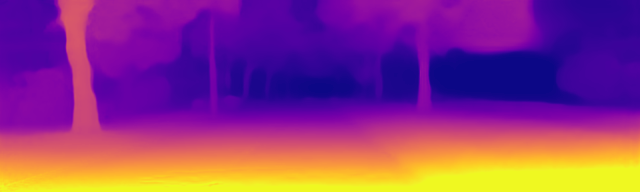}};
    \advance\vY by \vdY
    \node at (\vX,\vY) {\Pa{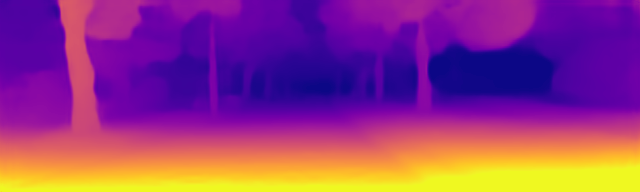}};

    \advance\vX by \vdX
    \vY = \vYi
    \node at (\vX,\vY) {\Pa{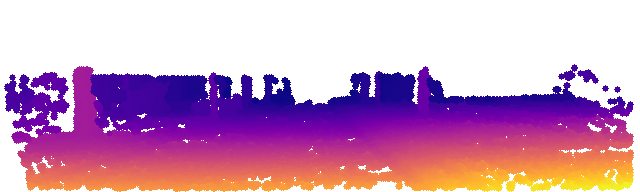}};

    \advance\vX by \vdXGT
    \advance\vY by \vdYGT
    \node[inner sep=0pt, font=\tiny] at (\vX,\vY) {GT};
    \advance\vX by -\vdXGT
    \advance\vY by -\vdYGT

    \advance\vY by \vdY
    \node at (\vX,\vY) {\Pa{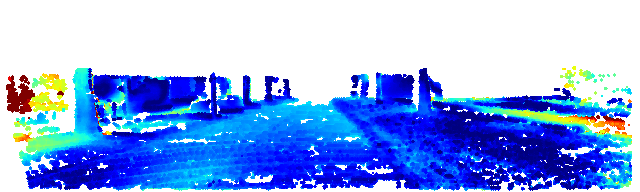}};

    \advance\vX by \vdXErr
    \advance\vY by \vdYErr
    \node[inner sep=0pt, font=\tiny] at (\vX,\vY) {Error map};
    \advance\vX by -\vdXErr
    \advance\vY by -\vdYErr

    \advance\vY by \vdY
    \node at (\vX,\vY) {\Pa{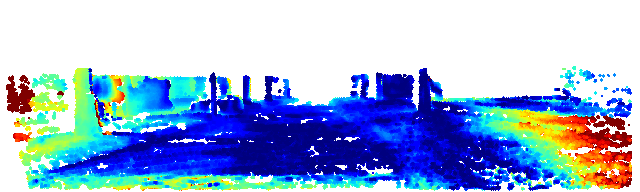}};
    \advance\vY by \vdY
    \node at (\vX,\vY) {\Pa{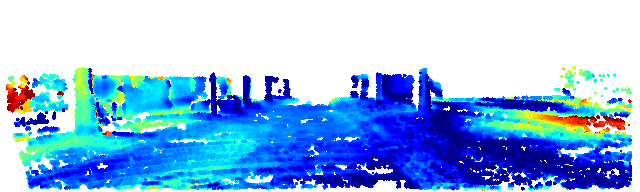}};
    \advance\vY by \vdY
    \node at (\vX,\vY) {\Pa{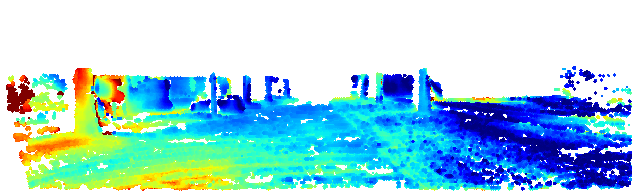}};

    \end{tikzpicture}
    \caption{Additional qualitative comparison on KITTI. We compare estimated depth maps of Monodepth2~\cite{Monodepth2} trained with FUMET to the ones of weakly-supervised methods: G2S~\cite{GPS}, DynaDepth~\cite{IMU}, and VADepth~\cite{VADepth}. In error maps, the larger depth errors are represented in red, and smaller ones are depicted in blue. The results show that FUMET can predict more accurate depth maps in various scenes, compared with the weakly-supervised methods.}
    \label{fig:KITTIQualitySupp}
\end{figure}

We provide additional results of FUMET in diverse scenes on the KITTI dataset. As shown in \cref{fig:KITTIQualitySupp}, FUMET can predict more accurate depth maps than the weakly-supervised methods, especially in road and object regions.

\end{document}